# Toward Generalizable Machine Learning Models in Speech, Language, and Hearing Sciences: Estimating Sample Size and Reducing Overfitting

Hamzeh Ghasemzadeh [a,b,c], Robert E. Hillman [a,b,d,e], Daryush D. Mehta [a,b,d,e]

[a] Center for Laryngeal Surgery and Voice Rehabilitation, Massachusetts General Hospital, Boston, MA, USA
[b] Department of Surgery, Harvard Medical School, Boston, MA, USA
[c] Department of Communicative Sciences and Disorders, Michigan State University, East Lansing, MI, USA
[d] Speech and Hearing Bioscience and Technology, Division of Medical Sciences, Harvard Medical School, Boston, MA, USA
[e] MGH Institute of Health Professions, Boston, MA, USA

Running Title: Power Analysis and Reducing Overfitting in Machine Learning

Email Address of the Authors:
Hamzeh Ghasemzadeh, email: hghasemzadeh@mgh.harvard.edu
Robert E. Hillman, email: hillman.robert@mgh.harvard.edu
Daryush D. Mehta, email: mehta.daryush@mgh.harvard.edu

Corresponding Author:
Hamzeh Ghasemzadeh
Center for Laryngeal Surgery and Voice Rehabilitation
Massachusetts General Hospital
One Bowdoin Square, 11th Floor
Boston, MA 02114

ABSTRACT
Purpose: Many studies using machine learning (ML) in speech, language, and hearing sciences rely upon cross-validations with single data splitting. This study's first purpose is to provide quantitative evidence that would incentivize researchers to instead use the more robust data splitting method of nested *k*-fold cross-validation. The second purpose is to present methods and MATLAB code to perform power analysis for ML-based analysis during the design of a study.

Method: First, the significant impact of different cross-validations on ML outcomes was demonstrated using real-world clinical data. Then, Monte Carlo simulations were used to quantify the interactions among the employed cross-validation method, the discriminative power of features, the dimensionality of the feature space, the dimensionality of the model, and the sample size. Four different cross-validation methods (single holdout, 10-fold, train-validation-test, and nested 10-fold) were compared based on the statistical power and confidence of the resulting ML models. Distributions of the null and alternative hypotheses were used to determine the minimum required sample size for obtaining a statistically significant outcome (5% significance) with 80% power. Statistical confidence of the model was defined as the probability of correct features being selected for inclusion in the final model.

Results: ML models generated based on the single holdout method had very low statistical power and confidence, leading to overestimation of classification accuracy. Conversely, the nested 10-fold cross-validation method resulted in the highest statistical confidence and power, while also providing an unbiased estimate of accuracy. The required sample size using the single holdout method could be 50% higher than what would be needed if nested *k*-fold cross-validation were used. Statistical confidence in the model based on nested *k*-fold cross-validation was as much as four times higher than the confidence obtained with the single holdout–based model. A computational model, MATLAB code, and lookup tables are provided to assist researchers with estimating the minimum sample size needed during study design.

Conclusion: The adoption of nested *k*-fold cross-validation is critical for unbiased and robust ML studies in the speech, language, and hearing sciences.

**INTRODUCTION**

Statistical analysis and methods are essential parts of the scientific discovery process, and they play important roles during the design of an experiment, its execution, and its final analysis. For example, statistical power analysis drives the decision about the required sample size during experimental design (Cohen, 2013; Jones et al., 2003). Additionally, inferential statistics allow for the study of a sample of a population whose findings can then generalize to the larger population (Field et al., 2012; Lowry, 2014). Statistical machine learning (ML) is a group of statistical tools and techniques that has gained a lot of attention in different branches of science, with particular application to fields that are relevant to this journal's audience, such as healthcare (Liu et al., 2019; Qayyum et al., 2020; Shen et al., 2021) and speech, language, and hearing sciences (Oleson et al., 2019). ML provides many advantages over conventional statistical analysis. For example, a single model can often incorporate features from different scales of measurement (nominal, ordinal, interval, and ratio). Also, the ability of ML to capture complex, high-dimensional, and non-linear interactions among different features can provide a significant advantage over conventional statistical methods. ML can be categorized into two main groups: supervised and unsupervised ML. In supervised ML, true labels of the data are known *a priori* (e.g., knowing which speech samples are recorded from individuals with Parkinson's disease or from neurotypical individuals). Supervised ML encompasses a wide variety of methods ranging from simple logistic regression (Menard, 2002) to deep learning (Liu et al., 2019) and ensemble learning (Ghasemzadeh, 2019b; Sagi & Rokach, 2018). In unsupervised ML, unlabeled data are used to discover underlying structures and patterns of the data (e.g., doing the same study but not knowing which voice data came from individuals with Parkinson's disease) (Theodoridis & Koutroumbas, 2009).

Studies using supervised ML in healthcare (including speech, language, and hearing sciences) can be categorized into two main groups depending on their aims: tool-developing and knowledge-developing. Tool-developing studies primarily use ML to "create a tool" for performing a task automatically. On the other hand, knowledge-developing studies (also referred to as ML-based science in the literature (Kapoor & Narayanan, 2022)) primarily use ML as a robust statistical analysis and data mining tool to gain knowledge and/or answer research questions about certain phenomena. Tool-developing studies closely resemble an engineering approach, and their aims could vary from automated diagnosis and evaluation to segmentation of a signal of interest, to solving an inverse problem. Examples of using ML to automatically differentiate healthy individuals from those with disorders (for automated screening/diagnosis and assessment of treatments) include the identification of voice disorders using voice samples (Arjmandi et al., 2011; Arjmandi & Pooyan, 2012; Ghasemzadeh et al., 2015), Parkinson's disease using speech recordings (Ghasemzadeh et al., 2022; Tsanas et al., 2012), amyolateral sclerosis using speech recordings (Ghasemzadeh & Searl, 2018; Vieira et al., 2022), septic infants from the audio signal produced by their cry (Matikolaie & Tadj, 2022), abnormal vocal folds from laryngoscopic images (Cho & Choi, 2020), intelligibility assessment of aphasic speech (Le et al., 2016), swallowing disorder from high-resolution manometry recordings (Mielens et al., 2012), screening for autism spectrum disorder (Crippa et al., 2015; Thabtah & Peebles, 2020), predicting the hearing outcome in patients with sudden sensorineural hearing loss (Bing et al., 2018; Uhm et al., 2021), and predicting language and communication skills of infants based on their speech-evoked electroencephalography data (Wong et al., 2021). Studies that have utilized ML for segmentation applications include automated segmentation of the glottis from high-speed videoendoscopy recordings (Kist et al., 2021), detection of high-speed videoendoscopy frames with partial occlusion of the vocal folds (Yousef et al., 2022), tongue segmentation from ultrasound images (Hamed Mozaffari & Lee, 2019), segmentation of videofluoroscopic recordings for swallowing studies (Donohue et al., 2021), speech classification for improving the performance of hearing aids (Bhat et al., 2020), and identification of the type of stuttering-related events from speech (Alharbi et al., 2020; Bayerl et al., 2022). Estimating parameters of the phonatory system from acoustic signals (Gómez et al., 2018; Ibarra et al., 2021; Zhang, 2020), measuring the distance between a laser-projection endoscope and the vocal folds (Ghasemzadeh et al., 2020), and compensating for non-linear image distortion in fiberoptic endoscopes and performing calibrated $mm$-measurements in laryngeal images (Ghasemzadeh et al., 2021) are some examples of solving an inverse problem in voice and speech science using ML.

On the other hand, in knowledge-developing studies, the main outcome is not the trained model itself, but rather the knowledge that has been gained. One example from this category would be the application of ML for understanding differences in phonatory functions between vocally healthy individuals and patients with phonotraumatic vocal hyperfunction using ambulatory recordings (Ghassemi et al., 2014; Mehta et al., 2015; Van Stan et al., 2020), or differences between mild and moderate severity of phonotrauma (Van Stan et al., 2023). Other examples would be the quantification of the information lost in the perceptual evaluation of voice due to inherent limitations of the human auditory system (Ghasemzadeh & Arjmandi, 2020), comparison between the information content of temporal and spectral features of connected speech for clinical evaluation of voice and speech (Ghasemzadeh et al., 2022), the optimum number of sensors for encoding articulatory information from electromagnetic articulography recordings (Wang et al., 2016), the effect of speech duration on classification uncertainty of persons with dementia (Ossewaarde et al., 2020), quantification of articulatory distinctiveness of different vowels and consonants (Wang et al., 2013), evaluating the relevance of different features in predicting hearing loss (Lenatti et al., 2022), and investigating the predictive power of different items of the Infant Monitoring Questionnaire on language outcomes (Armstrong et al., 2018).

**The Problem of Overfitting in Machine Learning**

Despite their growing popularity in the speech, language, and hearing sciences, ML models are susceptible to overfitting, which refers to models performing very well on one dataset but poorly on similar, but independent, datasets. One contributing factor to overfitting is that the decision boundary of a model is determined during a training process, and the boundary could be complex, non-linear, and in a high-dimensional space on the training dataset. The second contributing factor could be due to improper implementation of model selection or hyperparameter optimization components, which may lead to data leakage (Kapoor & Narayanan, 2022) and overestimation of model performance. Model selection refers to the process of choosing the best-performing model out of many candidate models. Feature selection is a common example of model selection in speech, language, and hearing sciences, which determines the best subset of features or measures that yield the highest performance. Hyperparameter optimization refers to the process of tunning parameters of an ML algorithm (e.g., parameters of a support vector machine) to get the maximum performance. Whereas the possibility of having a high-dimensional, complex, non-linear decision boundary and model selection account for many of the primary advantages of ML over conventional statistical analysis, they raise the possibility of overfitting.

Proper utilization of an ML step termed cross-validation can both alleviate overfitting (Hawkins, 2004) and estimate the generalizability of the trained model (Theodoridis & Koutroumbas, 2009). Cross-validation is a method that uses some portion of the data called the training set for training the model and then evaluates the performance of the trained model on the remainder of the data called the testing set. Different cross-validation approaches are explained later in the Cross-validation methods section of the paper. Reviewing the existing literature that used ML in the speech, language, and hearing sciences showed some important trends. For example, many of these investigations have employed simple cross-validation techniques only containing a single training set (see, for example, (Ibarra et al., 2021; Van Stan et al., 2020; Vieira et al., 2022; Yousef et al., 2022; Zhang, 2020)). Another significant observation was the sample size. While very large public datasets (several thousand to several hundred thousand samples) are available for machine vision (Huang et al., 2008) and speech processing (Panayotov et al., 2015), studies from the speech, language, and hearing fields are associated with small sample sizes (often a few tens of samples). Small sample sizes are especially susceptible to overfitting, with extra caution required for the proper application of cross-validation. This is especially true if the ML processing pipeline includes model selection components.

The problem of overestimation of ML performance in the presence of a model selection component has already been investigated in several studies. For example, two recent studies have reported very alarming findings in the application of ML in speech, language, and hearing literature (Berisha et al., 2021; Vabalas et al., 2019). A recent study surveyed the literature on the application of ML in autism and found a strong negative correlation between sample size and reported accuracy (Vabalas et al., 2019). Similar findings have been reported on speech-based classifications of individuals with Alzheimer's disease (Berisha et al., 2021) and for individuals with different forms of cognitive impairments (Berisha et al., 2021). These observed trends are at odds with theoretical expectations and a strong indication of widespread flawed evaluation of ML in the literature, indicating the possibility of a widespread overfitting and overestimation of the performance. A recent study has described this situation as a "crisis in ML-based science" (Kapoor & Narayanan, 2022) that was mainly attributed to data leakage, which often refers to the existence of some overlap (sometimes very subtle) between training and testing sets and not having "truly" independent test set(s). Application of cross-validation without a "true" test set in the presence of a model selection component is a common culprit for data leakage. Prior studies have shown that cross-validations with independent test sets are robust to overfitting and can provide an unbiased estimate of ML performance (Krstajic et al., 2014; Parvandeh et al., 2020; Varoquaux et al., 2017). Simulated data based on Gaussian distributions have been used to study the effect of different cross-validations in the presence of feature selection and model parameter optimization (Vabalas et al., 2019).

**Power Analysis in Machine Learning**

When evaluating different cross-validation methods to reduce overfitting, prior literature has quantified their effect on ML performance. However, power analysis of ML and the effect of different cross-validations on the required sample size have not been addressed. For example, a recent meta-analysis of ML in medical imaging applications found that, out of 167 papers, only four papers included some procedures for determining sample size (Balki et al., 2019). In contrast, power analysis before conducting a study is a common practice in science and a ubiquitous component of grant applications to the National Institutes of Health. These highlight the importance of this missing piece in knowledge-developing ML studies. In conventional statistical tests, power analysis allows researchers to estimate the minimum sample size required for achieving a target statistical power. This analysis is often done based on effect sizes reported in the literature. However, with ML, power analysis can be defined in multiple ways. This study divides them into two broad categories: "the *required* sample size" and "the *recommended* sample size." The *required* sample size follows the conventional definition, and it is the "minimum" sample size required to have enough statistical power to reject the null hypothesis. In contrast, the *recommended* sample size is based on reaching a target performance for the ML model. A common target performance in the ML community has been the ML accuracy reaching a plateau, after which only marginal improvement is observed for increasing the sample size beyond that point (Figueroa et al., 2012).

A comprehensive review of the "recommended sample size" approach was presented in a recent study (Viering & Loog, 2022). This definition of recommended sample size and the adopted methodology in Figueroa et al., 2012 and its referenced studies neglect the effect size reported in the literature, which is a significant source of information and a very powerful aspect of power analysis in conventional statistical analysis. Also, these methods are more appropriate for active learning paradigms (Figueroa et al., 2012), where ML performance gets evaluated continuously as batches of new samples are added to the dataset (often through labeling the data and not study participant recruitment), and hence not applicable to estimating the required sample size during the design of a study. The initial recommended dataset for these methods is also very large (100 to 200 samples (Figueroa et al., 2012)), and the final sample size could be much larger than this number. Such sample sizes are practical in tool-developing studies, where often the data just need to be labeled. However, knowledge-developing studies involve experimental design and active recruitment of participants, and such high sample sizes would often be impractical.

In summary, the prior sample size estimation studies were focused on the recommended sample size and not the required sample size. Additionally, the prior works did not consider the effect of different cross-validations on the sample size. The current study incorporates the traditional definition of power analysis into the ML framework. This new approach would allow researchers to use the effect sizes reported in the literature for power analysis during the design of an ML study. Finally, an alternative definition of recommended sample size based on the effect sizes reported in the literature is presented in this study that relates to the optimality of feature selection and its ability to find the correct subset of features. We will argue why this alternative definition of recommended sample size is important and more relevant for knowledge-developing ML studies.

**Study Aim and Research Questions**

Optimization (determining the value of a parameter given a particular desired outcome) is an integral component of ML models. Two different types of optimizations can be differentiated in the context of ML. First, determining the optimum decision boundary between different classes which is part of the training process and, hence, present in any ML implementation. Second, model selection which may include feature selection, hyperparameter optimization, optimization of the architecture of a (deep or shallow) neural network, selecting between different classification algorithms, etc. The focus of this study is on ML implementations that also include a model selection component. Also, we paid specific attention to feature selection in this study. However, the general findings should apply to other model selection scenarios.

The main aim of this study was to show that the choice of cross-validation method has a profound impact on the statistical power of ML and to provide a strong motivation for researchers in speech, language, and hearing sciences and other disciplines to migrate from the commonly used single holdout and train-validation-test cross-validations to more robust and powerful methods. Specifically, this study quantified the statistical relationship among different cross-validation methods, the minimum required sample size, the number of extracted features, the number of selected features, and the discriminative power of selected features. The outcome of this study could be used to determine the required and recommended sample sizes of an ML study during its experimental design, to select the appropriate dimensionality of the feature space for an existing dataset, and to help scientists with the proper implementation of their ML processing pipeline to achieve generalizable ML models. Last but not least, the availability of power analysis method for ML could lead to a wider application of ML in clinical science where researchers may have previously thought that the sample sizes for their studies could be too small to apply ML. To pursue this aim, four research questions were asked in this study.

**Q1**: How is the statistical power of an ML model affected by the employed cross-validation method?
**Q2**: How is the statistical confidence in an ML model affected by the employed cross-validation method?
**Q3**: What is the minimum required sample size to get a statistically significant outcome from an ML model with conventional power requirements (5% significance, 80% power [$\alpha=0.05$, $1-\beta=0.8$])?
**Q4**: What is the recommended sample size for achieving a target statistical confidence of the final ML model?

This paper is organized as follows. First, the cross-validation methods that were investigated in this study are presented. Then, the performance of ML on a clinical dataset is reported to demonstrate the profound effect of different cross-validations on the statistical properties of the trained ML models. The implications of these differences and the rationale for using simulated data for the rest of the paper are also presented in this section. The framework for generating simulated data and statistical evaluation of ML models are presented next. The comparison of cross-validation approaches using simulated data section is devoted to reporting the outcomes of three different experiments designed to answer the four research questions. Experiments 1 and 2 answer research questions Q1 and Q2, respectively. Experiment 3 was conducted to answer research questions Q3 and Q4. The results of the experiments are put into the larger context in the Discussion section, where more general implications of this work are presented, along with limitations of this study, generalization of its findings, and directions for future studies.

**CROSS-VALIDATION METHODS**

Feature selection is an optimization problem, and the cost function (the evaluation metric that is to be maximized) is the accuracy of the classifier (Theodoridis & Koutroumbas, 2009). However, depending on which cross-validation is implemented, the cost function is computed differently. Therefore, our main hypothesis was that the choice of cross-validation would significantly affect the statistical characteristics (average accuracy, standard deviation (std) accuracy, and confidence) of the final, optimized ML model. Although cross-validation comes in several different flavors, they can be categorized into two main types: two-split and three-split cross-validations. In the two-split methods, the dataset is divided into (a series of) two separate partitions of training and testing sets, whereas in the three-split methods, the dataset is divided into (a series of) three separate partitions of training, validation, and testing sets. This paper compares two two-split methods and two three-split methods of cross-validation.

*Single holdout method*

The single holdout approach is a two-split method and the simplest cross-validation approach. In this approach, based on a target splitting ratio, the dataset is randomly divided into two disjoint sets: a training set and a testing set. The training set is used for creating the model. The accuracy of the classifier is computed by applying the trained model to the testing set. Therefore, the cost function of feature selection is the accuracy of the testing set. This means the feature set that maximizes the testing accuracy will be selected for the final, optimized ML model. The main advantages of this approach are its implementation simplicity and low computational complexity. Figure 1(A) shows a schematic of this approach. In this study, we used a stratified version of the single holdout cross-validation to ensure that both training and testing sets had the same proportion of samples from each class, with a splitting ratio of 70% (training) to 30% (testing).

### *k-fold cross-validation*

The *k*-fold cross-validation approach is also a two-split method, where the dataset is divided randomly and evenly into *k* disjoint sets. Each set contains $2n/k$ samples (*n* is the sample size per each class) and is considered the testing set for that fold. All the remaining samples constitute the training samples for that fold. Similar to the single holdout, the training set from each fold is used for training the model, and then that model is applied to the corresponding testing set to get an estimate of model performance. However, in contrast with the single holdout method, the training/testing process is performed multiple (*k*) times. The cost function of feature selection is the average accuracy over the *k* different testing sets. This means that the feature set that maximizes the average testing accuracy will be selected for the final, optimized ML model.

The main advantages of this approach are that all samples contribute to the training and testing of the ML model, and the robustness of model performance can be obtained by computing the std of the *k* testing accuracies. Figure 1(B) shows a schematic of this approach with $k = 3$ folds for visualization purposes. In this study, we used stratified 10-fold cross-validation. The selection of $k = 10$ was motivated by its common use in the literature and a recent study supporting the suitability of this selection (Marcot & Hanea, 2021).

### *Train-validation-test cross-validation*

The train-validation-test approach is a three-split method, where the dataset is first split into two disjoint sets termed training-validation and testing sets. The training-validation set is then further split into two disjoint sets of training and validation sets using single holdout, *k*-fold cross-validation, or a different method. An ML model is trained using the training set and is then applied to the associated validation set to obtain an estimate of model performance. However, unlike the previous two-split methods, the cost function of feature selection is either validation accuracy (in the case of single holdout) or average validation accuracy over multiple splits (in the case of *k*-fold cross-validation). This means that the feature set that maximizes accuracy across validation set(s) will be selected for the final, optimized model. The selected features are then used to train the final model using both training and validation samples. Critically, the final model is then applied to the testing samples to obtain a single value of model performance on a disjoint testing set.

The main advantage of this approach is that the testing set is completely independent of the model optimization process (i.e., feature selection in our case). Figure 1(C) shows a schematic of this approach with $k = 3$ folds within the training-validation set for visualization purposes. In this study, we used a stratified version and reserved 15% of the samples for the testing set. The remaining 85% of samples were divided into training and validation sets using 10-fold cross-validation.

### *Nested k-fold cross-validation*

The nested *k*-fold cross-validation approach is also a three-split method, and is, basically, a *k*-fold (inner) cross-validation embedded inside another *k*-fold (outer) cross-validation. The nested *k*-fold cross-validation method thus consists of an outer *k*-fold cross-validation that divides the data into *k* different folds, with each outer fold consisting of training-validation and testing sets. Each training-validation set undergoes a secondary (inner) *k*-fold data splitting to create training and validation sets. In this sense, the testing sets of the outer folds represent multiple independent testing sets, the testing sets of the inner folds are intermediate validation sets, and the training sets of the inner folds are the training sets. This means that, unlike the previous three methods, feature selection will be executed *k* different times (one per each outer fold) and hence we will have *k* different estimates for the best feature set. In each outer fold, the feature set that maximizes the average accuracy over the *k* different validation sets (one for each inner fold) is selected as a candidate best feature set. Also, in contrast with the previous cross-validation methods, a consensus criterion is needed to select the best feature set among the *k* candidate feature sets (one from each outer fold) (Parvandeh et al., 2020). Figure 1(D) shows a schematic of nested *k*-fold cross-validation with three outer folds and three inner folds per outer fold. In this study, we employed $k = 10$ and selected the feature set that had the highest probability of joint occurrence among the *k* different best feature sets (one per outer fold).

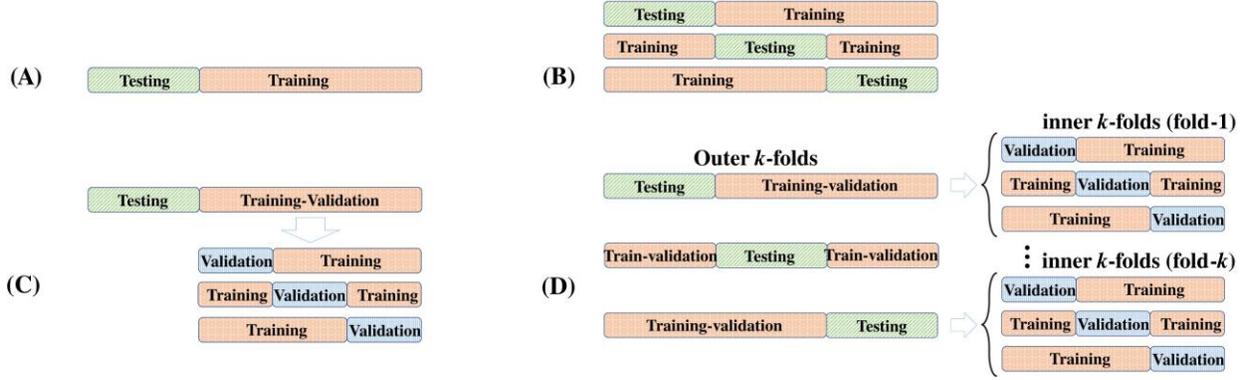

Figure 1. Visualization of data splitting for the four cross-validation methods studied: (A) single holdout, (B) $k$-fold ($k = 3$), (C) train-validation-test, (D) nested $k$-fold cross-validation ($k = 3$).

**COMPARING CROSS-VALIDATION METHODS IN A REAL-WORLD CLINICAL EXAMPLE**

The significant impact of different cross-validations is demonstrated here using ambulatory voice monitoring data. A primary focus of our research group has been determining the real-world vocal behavior/function associated with hyperfunctional voice disorders using ambulatory voice monitoring technology (Ghassemi et al., 2014; Mehta et al., 2015; Van Stan et al., 2020). To that end, six different voice measures capturing different aspects of phonation were extracted from the ambulatory data of patients with phonotraumatic vocal hyperfunction (PVH) and vocally healthy matched controls. The extracted measures included fundamental frequency ($f_o$), cepstral peak prominence (CPP), sound pressure level (SPL), the difference between amplitudes of the first and second harmonics of the signal (H1-H2), smoothed duration of phonatory segments (voicing duration), and duration of non-phonatory segments (resting duration). The smoothed duration of phonatory segments was estimated from the outcome of the voice activity detector by connecting contiguous voiced segments that were separated by half a second or less of non-phonatory activity. No smoothing was applied to the durations of the non-phonatory segments. The methodology of processing the ambulatory data and the definition of these measures have been reported in an earlier study (Mehta et al., 2015). Similar to prior work (Ghassemi et al., 2014; Mehta et al., 2015; Van Stan et al., 2020), the day-long distribution of each measure was constructed and eight different distributional characteristics (mean, median, std, interquartile range, $5^{th}$ percentile, $95^{th}$ percentile, skewness, and kurtosis) were extracted. The final classification features were computed as distributional characteristics of each participant, averaged over all recording days, meaning each participant contributed only one data point to the dataset. Therefore, the dimensionality of the feature space was 48 (6 voice measures with 8 distributional characteristics). Our dataset included recordings from 153 females with PVH and 136 female controls. The dataset was balanced by selecting 136 patients randomly. To make the findings robust to the random selection of patients and random splitting of the data, each analysis was repeated 500 times.

First, the average and std of testing accuracy for different numbers of selected features and different cross-validation methods are presented in Figure 2(A). The results showed dissimilar trends for different cross-validation. Specifically, two-split cross-validation methods (single holdout, 10-fold) yielded higher average accuracy compared to the three-split cross-validation methods (train-validation-test, nested 10-fold), with single holdout tending to yield the highest average accuracy for each number of selected features. Another significant observation was the impact of different cross-validation methods on the variability of estimated testing accuracy. Specifically, testing accuracies of the methods with a single testing set (single holdout, train-validation-test) were quite sensitive to the random splitting of the data (larger error bars), whereas methods with multiple testing sets (10-fold, nested 10-fold) were quite robust (smaller error bars). In summary, the selection of cross-validation method has a significant impact on the distributional characteristics of the estimated performance of ML. The implications of these observations on the generalizability of the findings and the statistical power of the model will be discussed at the end of this section.

Second, the effect of changing the sample size from 25 pairs to 125 pairs on the testing accuracy of the model with four selected features was evaluated. Similar to the previous analysis, the participants were selected randomly, and each analysis was repeated 500 times. The average and std of testing accuracy for different sample sizes are presented in Figure 2(B). Interestingly, the estimated performances of cross-validations with two splits (single holdout, 10-fold) declined as the sample size was increased, which doesn't follow the theoretical expectation. Conversely, testing accuracies of cross-validations with three splits (train-validation-test, nested 10-fold) increased as the sample size was increased. The implications of these observations on the generalizability of the findings and the statistical power of the model will be discussed at the end of this section.

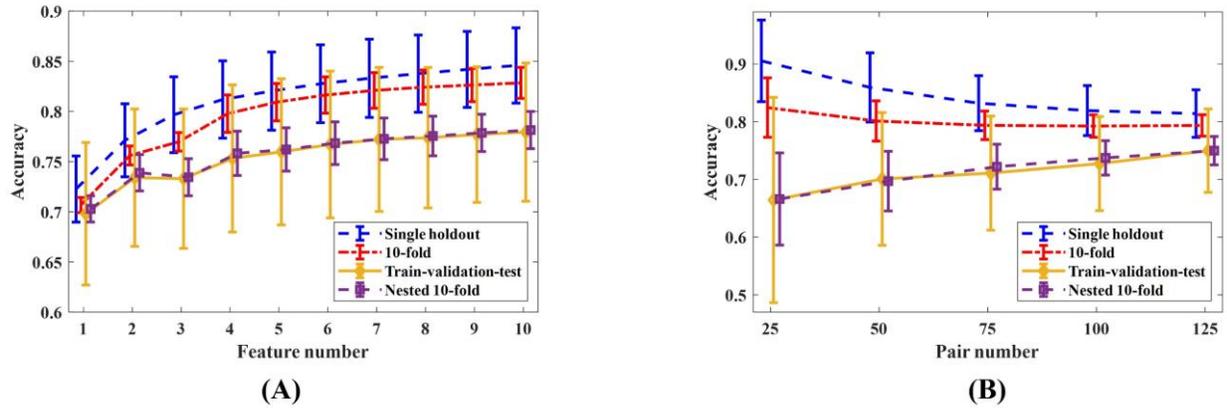

Figure 2. (A): The effect of different cross-validations on average and std of testing accuracy for different numbers of selected features, (B): the effect of sample size on testing accuracy estimated based on different cross-validations.

Third, as we discussed earlier, the knowledge that is gained (vs. the trained model itself) is the primary objective of knowledge-developing ML studies. Looking at features that are selected and the orders in which they get selected is an important source of acquiring such knowledge. For example, in this clinical case study of ambulatory voice monitoring, each of the six voice measures relates to a different aspect of phonation. The finding that a distributional characteristic of H1-H2 is the best feature versus a distributional characteristic of $f_o$ would have different implications regarding the etiology and/or compensatory behavior associated with PVH. To demonstrate the effect of different cross-validation methods on the outcome of feature selection, the probability of different features being selected at different steps of feature selection was computed from the 500 repetitions of the first analysis of this section. In this fashion, the first feature was the feature that provided the highest classification accuracy. The second feature was the feature (among the remaining ones) that provided the highest improvement in classification accuracy. The third and the fourth features were defined in a similar way.

Figure 3 shows the results. According to Figure 3(A), the outcome of feature selection for the single holdout method was very unstable (low selection probabilities), meaning the outcome of feature selection was very sensitive to how the data were split into training and testing sets. That is, depending on the samples that were in the training set, different features (and even from different voice measures) were selected as the most informative (i.e., the highest classification accuracy) descriptors of PVH, which indicates low generalizability for the trained model. The situation for the train-validation-test method was better, where the first and second features were consistently selected (high selection probabilities). However, the third and fourth features had low selection probabilities. Lastly, 10-fold and nested 10-fold cross-validation methods produced models that were much more consistent, with the highest consistency of feature selection exhibited by nested 10-fold cross-validation.

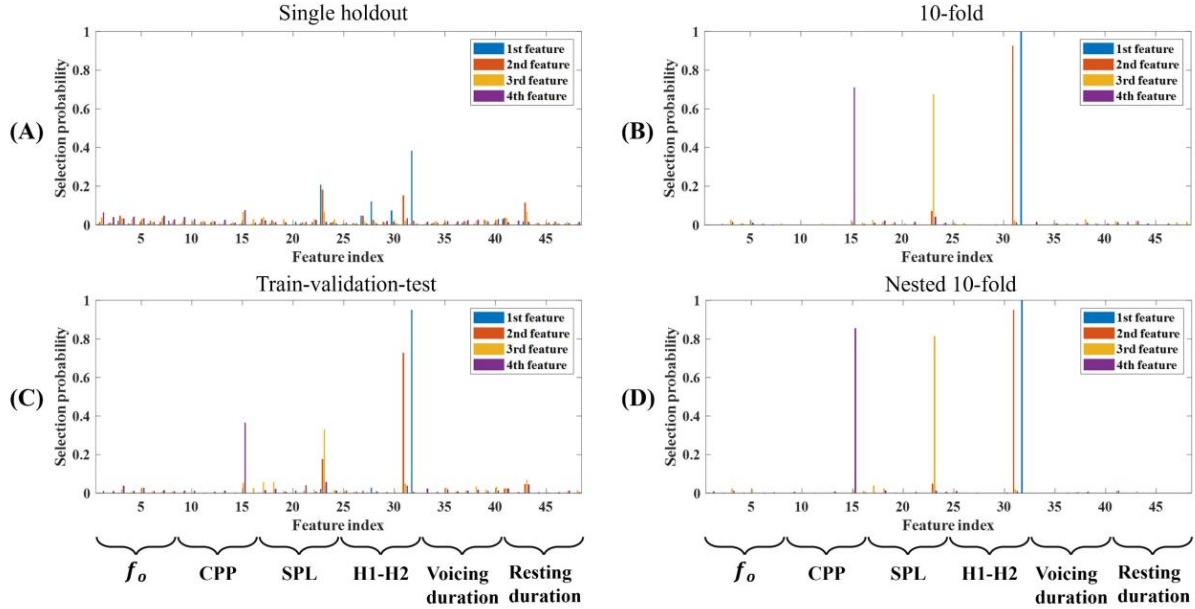

Figure 3. A comparison of statistical confidence of four cross-validation methods: (A) single holdout, (B) 10-fold, (C) train-validation-test, (D) nested 10-fold. The horizontal axes represent the 48 dimensions in the feature space (the 6 voice measures listed under the panels with 8 distributional characteristics per voice measure). Within each voice measure category on the plots, the distributional characteristics, in order, are the mean, median, std, interquartile range, 5th percentile, 95th percentile, skewness, and kurtosis.

These three analyses showed the impact of different cross-validation methods on the ML outcomes, highlighting the importance of employing appropriate cross-validation for ML in healthcare. Based on the result of Figure 2(A), the average and std of the distribution of classification accuracy depended on how cross-validation was implemented. Figure 2(B) showed that increasing the sample size had dissimilar effects on the distribution of the accuracy of the classifier for different cross-validations. The statistical power of ML directly relates to the distribution of its accuracy; therefore, we can expect different statistical power for different implementations of cross-validations. Now the first question (research question Q1) is which method has the highest statistical power (and hence requires the smallest sample size). To answer this question, we need to systematically change the discriminative power of features, the sample size of the dataset, the dimensionality of feature space, etc. Also, we need to have thousands of independent examples of each case so we can generate the distribution of the performance that is required for power analysis. Therefore, simulated data were used for the rest of this paper.

Based on the results of Figure 3, the statistical confidence of the model varied significantly among the different cross-validation methods. The next question (research question Q2) is which method gives us the correct answer. With real data, we could not be sure which features are the most discriminative ones. Also, we cannot systematically change the discriminative power of features, the sample size of the dataset, the dimensionality of feature space, etc., and determine the recommended sample size for meeting a target model confidence. In contrast, with simulated data, we know the underlying distribution of each feature and, hence, which features should be selected. Also, we can change the parameters of the problem systematically. As a final note, more than 20 different voice measures have been presented in prior ambulatory voice monitoring studies (Cortés et al., 2018; Mehta et al., 2015; Van Stan et al., 2020). Additionally, up to 10 different distributional characteristics have been used in prior studies. This would add up to more than 200 possible features. A practical follow-up question would be, how many features can be extracted for a given sample size to have statistical confidence in the findings? The following simulation study will present methodologies to answer these important and practical questions.

**STATISTICAL FRAMEWORK FOR MACHINE LEARNING MODEL EVALUATION**

This study assumed a balanced two-class classification problem (e.g., detecting the presence of a disorder in an equal number of patients and controls in a clinical application) with a sample size of $2n$ participants ($n$ per class). A feature selection component was also included in the ML processing pipeline. Therefore, this study assumed a total of $m$ independent computed features, of which only $l$ were discriminative. All analyses were executed on a server running a Windows 2019 Server, with two Xenon Gold 6230 CPUs, and 768 GB of RAM. We used MATLAB R2021b for our analyses.

Table 1. Definition of symbols and parameters used in this study.

| Symbols | Definition |
|---|---|
| $H_0$ | The null hypothesis |
| $H_a$ | The alternative hypothesis |
| $\alpha$ | Probability of type I error |
| $\beta$ | Probability of type II error ($1-\beta$ is called power) |
| $\mathcal{N}_m$ | $m$-dimensional Gaussian distribution |
| P | Samples from the positive class |
| N | Samples from the negative class |
| $\boldsymbol{\mu}_P$ | An $m \times 1$ vector of means of $\mathcal{N}_m$ for positive samples |
| $\boldsymbol{\Sigma}_P$ | An $m \times m$ matrix storing the covariance of $\mathcal{N}_m$ for positive samples |
| $\mathbf{I}_m$ | Identity matrix with the size of $m \times m$ |
| $m$ | The dimensionality of the feature space |
| $n$ | Sample size per each class in a balanced dataset |
| $l$ | The number of discriminative (i.e., selected) features |
| $D$ | Cohen's D effect size of discriminative features |
| $k$ | The number of folds in $k$-fold and nested $k$-fold cross-validation |
| $C_{l,d}$ | Statistical confidence of the model (the probability of correct selection of $d$ features out of $l$) |
| $\theta_i$ | Parameters of fitted exponential curves |
| r | Linear correlation coefficient |
| p | $p$-value |
| $n_r$ | The required sample size to get a statistically significant outcome with conventional power requirements ($\alpha$=0.05, $1-\beta$=0.8) |
| $\gamma_{Db}$ | The ratio of the number of positive to the number of negative samples in an unbalanced dataset |
| $\gamma_D$ | The ratio of the larger Cohen's D to the smaller one if selected features had different discriminative powers. |

Given the aims of this study, features of positive (e.g., patient group) and negative (e.g., healthy control group) classes were generated from a series of multivariate Gaussian distributions. Additionally, to make the simulation tractable, we assumed all discriminative features to have the same discriminatory power of $D$ (Cohen's $D$ effect size metric (Cohen, 2013)). Let $\mathcal{N}_m$ denotes an $m$-dimensional Gaussian distribution with $\boldsymbol{\mu}_P$ ($\boldsymbol{\mu}_N$) being an $m \times 1$ vector of means and $\boldsymbol{\Sigma}_P$ ($\boldsymbol{\Sigma}_N$) being an $m \times m$ matrix storing the covariance of the positive (negative) samples, where $m$ is the number of features extracted. Equations 1 and 2 show the formula for data generation, where $\boldsymbol{X}_P$ and $\boldsymbol{X}_N$ are $n \times m$ matrices storing $n$ samples from the positive class and $n$ samples from the negative class, respectively.

$$\boldsymbol{X}_P \sim \mathcal{N}_m(\boldsymbol{\mu}_P, \boldsymbol{\Sigma}_P) \qquad (1)$$
$$\boldsymbol{X}_N \sim \mathcal{N}_m(\boldsymbol{\mu}_N, \boldsymbol{\Sigma}_N) \qquad (2)$$

For the negative class, we set $\boldsymbol{\mu}_N = [0, 0, ..., 0]^T$, and for the positive class we used $\boldsymbol{\mu}_P = [D, ..., D, 0, ..., 0]^T$, with the first $l$ elements of $\boldsymbol{\mu}_P$ equal to $D$ and the remaining $m-l$ elements equal to 0. Additionally, we used $\boldsymbol{\Sigma}_N = \boldsymbol{\Sigma}_P = \mathbf{I}_m$, where $\mathbf{I}_m$ is an identity matrix with the size of $m \times m$.

Comparing the definition of the $\boldsymbol{X}_P$ and $\boldsymbol{X}_N$ shows that only the first $l$ features of the positive and negative classes had different distributions and hence were discriminative, and the remaining $m-l$ features were irrelevant. The goal is then to find those $l$ features and then train a classifier on them. This goal is an optimization problem and can be broken into two separate components: (1) selecting a proper cost function and (2) adopting a strategy for searching the solution space. This study adopted a wrapper-forward feature selection method due to its high performance and low computational cost (Ghasemzadeh, 2019a). The cost function of a wrapper method is the performance of a classifier. This means the subset of features that maximizes the classification performance gets selected. Forward feature selection refers to the search strategy. It is an iterative greedy search algorithm that starts with an empty selected set. At each stage, the remaining features are sequentially added to the selected set. The feature that leads to the maximum improvement of the cost function is added to the selected set. The process is repeated until the addition of new features doesn't improve the cost function. Finally, logistic regression was used as the classifier. This was primarily motivated by its widespread usage in speech, language, and hearing sciences, the simplicity of its algorithm that offers interpretability and produces a closed-form model, and the underlying distributions of the positive and negative classes. Specifically, the positive and negative samples were generated based on multivariate Gaussian distributions with the same covariance but different means; therefore, the true decision boundary should be a hyperplane. Logistic regression implements such a decision boundary and hence would be optimal for our problem. See Table 1 for a summary of symbols and parameters used in this study.

**Statistical evaluation criteria**

To answer Q1 and Q3, the null ($H_0$) and alternative ($H_a$) hypotheses need to be formed. There is an implicit assumption in the application of supervised ML that the extracted features have better discriminative power compared to random features. This assumption can be used to formulate $H_0$ and $H_a$ for a supervised ML problem.

**$H_a$:** Extracted features will perform significantly better than random features.

To formally test this directional hypothesis, we need to construct the distributions of $H_a$ (i.e., the performance of the extracted features) and $H_0$ (i.e., the performance of random features). For this study, the distribution of $H_a$ can be estimated using Monte Carlo simulations by generating a series of $\boldsymbol{X_P}$ (Eq. 1) and $\boldsymbol{X_N}$ (Eq. 2) and passing them through our ML processing pipelines. The distribution of $H_0$ can be estimated similarly by setting D = 0. Once distributions of $H_0$ and $H_a$ are estimated, they can be used for quantitative evaluations of different ML processing pipelines (e.g., different cross-validation approaches). This study has used the following evaluation criteria.

*Distributional characteristics of $H_0$*

The distributional characteristics (mean, std, confidence interval) of $H_0$ are evaluation criteria that provide valuable insights into the statistical properties of different ML processing pipelines. For example, the mean accuracy of $H_0$ for a balanced two-class problem should be 0.5 (on a scale of 0 to 1), and any gain over that number is often interpreted as the discriminative power of the extracted features. This interpretation is based on an implicit assumption that the implemented ML processing pipeline is not biased. The distribution of $H_0$ would allow us to evaluate if such a bias exists and then quantify it. Additionally, based on a specific statistical significance level ($\alpha$), we can find the confidence interval (CI) of $H_0$. Investigating $H_a$ shows that the alternative hypothesis is directional. Therefore, the CI of $H_0$ would be the range of values covering the minimum of the $H_0$ distribution to its $(1-\alpha)\times 100$ percentile. Assuming a normal distribution, the mean and variance of $H_0$ contribute to its CI; therefore, the std of the $H_0$ distribution was another evaluation criterion for comparing different cross-validation approaches.

*Statistical power and the required sample size*

Statistical power is the probability of a test producing true positive results, i.e., the probability of detecting the presence of an effect if effect exists (Field et al., 2012). The distribution of $H_a$ and the target probability of type II error ($\beta$) are the factors determining the power of a test, where power is defined as $1-\beta$. Statistical power is closely related to the required sample size for a study to detect a significant effect with a reasonable probability. The required sample size depends on the $H_0$ and $H_a$ distributions and the selected values for $\alpha$ and $\beta$. To determine the required sample size for different cross-validation approaches, the distributions of $H_a$ and $H_0$ were estimated for different values of $n$ (Eqs. 1 and 2). Then, for a given value of $\alpha$, the upper bounds of the $H_0$ CI, and, for a given value of $\beta$, the lower bounds of the $H_a$ CI were estimated and plotted. The point of intersection between the two plots corresponds to the minimum required sample size. Figure 4 illustrates this for the commonly used values of $\alpha = 0.05$ and $\beta = 0.2$ (power of 0.8), which were used throughout this study.

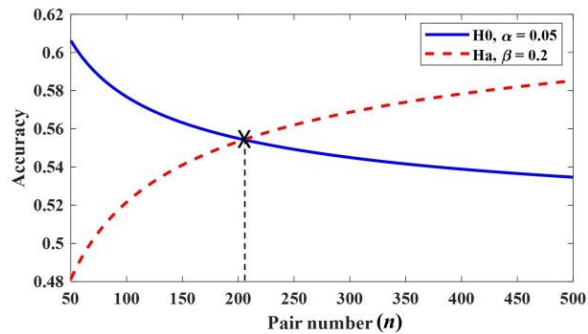

Figure 4. Empirical estimation of the minimum required sample size as the value of $n$ at the intersection between the upper bound of the $H_0$ confidence interval (given $\alpha = 0.05$) and lower bound of the $H_a$ confidence interval (given $\beta = 0.2$).

*Statistical confidence*

ML is a statistical tool, and hence, there is a chance that the outcome of an ML model provides incorrect results. Such inaccuracies are often quantified using performance metrics such as accuracy, sensitivity, and specificity. While it is well understood that the classifier outcome could be incorrect, the correctness of the model itself has received less attention. The correctness of the model can be defined in terms of the features that have been included (i.e., selected) in the model, as well as the coefficients (e.g., feature weights) of the model. The focus of this study is on the correctness of selected features.

Statistical confidence of an ML model is defined as the probability that correct features were selected and hence included in the ML model. In practice, statistical confidence is the probability of correct features being included in the final model if we repeat our experiment many times with different samples drawn from the population. This concept is especially important as it will have direct consequences on the interpretation that we make about models and the knowledge that we gain from them. As an example, let us consider the application of ML for understanding differences in vocal function and behavior between patients with a voice disorder and vocally healthy individuals. Depending on which features are being selected, we may attribute and rank different aspects of vocal function and/or behavior as the most likely contributors to causing and/or maintaining the disorder. Now, if the model has low confidence and other researchers try to replicate the study, there is a high probability that they will find very different results. This means, our conclusions about factors associated with the disorder could be unreliable, and, hence, we may target incorrect (or at least suboptimal) aspects of phonation during efforts to prevent and treat the disorder. Statistical confidence is a metric that measures such an attribute. Referring to the definition of $X_P$ and $X_N$ (Eqs. 1,2), we know which features should be included in the final mode, hence, the effect of different parameters and cross-validation approaches on the statistical confidence of the model can be quantified. The symbol $C_{l,d}$ will be used to represent the statistical confidence of the model for different numbers of selected features. Specifically, $C_{l,d}$ is the probability that, among the $l$ selected features in an ML model, $d$ of the features were correctly selected.

## COMPARISON OF CROSS-VALIDATION APPROACHES USING SIMULATED DATA
### Experiment 1: Distributional characteristics of $H_0$ and statistical power

Experiment 1 addressed research question Q1 ("How is the statistical power of an ML model affected by the employed cross-validation method?") by investigating the effect of different parameters on obtaining statistically significant outcomes from a supervised ML model. First, the effects of the four cross-validation methods and different sample sizes on the distribution of $H_0$ were investigated. Values of $m = 20$ (20 extracted features), $l = 2$ (2 selected features), and $D = 0$ were used. Figure 5 shows the results of running each experiment 5000 times, with each experiment defined by a given cross-validation method and sample size per class (50 to 500 samples, in increments of 50). The plots show the outcome of fitting a series of two-term exponential functions on the result of each cross-validation. The form of the exponential function was:

$$y = \theta_1 e^{\theta_2 x} + \theta_3 e^{\theta_4 x}, \qquad (3)$$

where $x$ and $y$ are the predictor and outcome variables, respectively, and $\theta_1, \theta_2, \theta_3, \theta_4$ are parameters of the exponential function.

In Figure 5(A), the mean accuracies of the single holdout and 10-fold cross-validations significantly deviate from the theoretically expected value of 0.5 for a balanced binary classification with irrelevant (non-discriminative) features. However, as the sample size increases, the mean accuracies of these two approaches start to converge toward the expected chance value of 0.5, indicating the accuracy that we saw at smaller sample sizes was due to overfitting and not learning the underlying patterns of the data (as there are no patterns for $H_0$). On the other hand, the average accuracies of the other two methods are quite close to the expected value of 0.5, even for small sample sizes. Comparing the mean of $H_0$ accuracy among different cross-validations clearly shows that the lack of an independent test set in the presence of an optimization task (e.g., feature selection, hyperparameter optimization, selection of the architecture of a neural network, etc.) leads to significant overestimation of the performance and should be avoided.

The whiskers of Figure 5(A) depict the standard deviation of the accuracy of different approaches across the 5000 experimental trials. The most significant observations are that the train-validation-test and 10-fold cross-validation methods had the highest and the lowest standard deviations, respectively. Also, we see a significant decrease in the value of dispersion as we move from the simple train-validation-test to the more sophisticated method of nested 10-fold cross-validation. Finally, the upper bound of the $H_0$ CI (one-sided, $\alpha = 0.05$) depends on both the mean and the dispersion of the $H_0$ distribution. Theoretically speaking, the upper bound increases with increases in the mean and dispersion of the $H_0$ distribution. This is also reflected in Figure 5(B). Based on this figure, for all approaches, the upper bound of $H_0$ decreases monotonically with an increase in the sample size. However, we see significant differences among the approaches, especially in lower sample sizes. For example, with 50 pairs, the single holdout method has a significant chance (more than 5%) to get accuracies as high as 76.7% even if the features are irrelevant (i.e., random features). However, this value drops to 62% if nested 10-fold cross-validation is used instead.

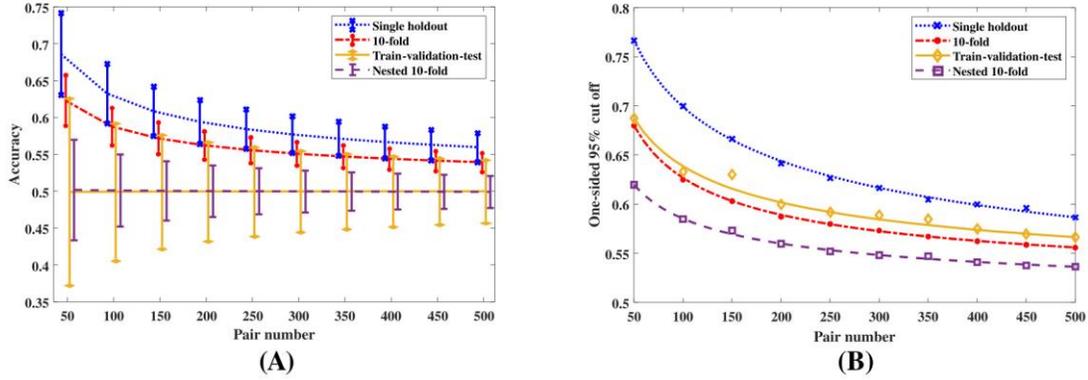

Figure 5. $H_0$ characteristics: (A) mean and standard deviation of $H_0$ accuracy (B) upper bound of $H_0$ CI (one-sided, $\alpha = 0.05$).

Second, the interaction effect between the dimensionality of the feature space ($m$) and different cross-validation methods was investigated. Figure 6 shows the results. Based on Figure 6(A), the bias (the deviation from the expected theoretical mean accuracy of 0.5) for single holdout and 10-fold cross-validation increases significantly as the number of extracted features increases, whereas methods with independent test sets (train-validation-test and nested $k$-fold) remain unbiased. Figure 6(B) represents the dependency of the upper bound of $H_0$ CI. Based on this figure, the upper bound of $H_0$ CI increases with the number of extracted features for the single holdout and 10-fold cross-validations, whereas this upper bound remains relatively constant for the methods with independent test sets.

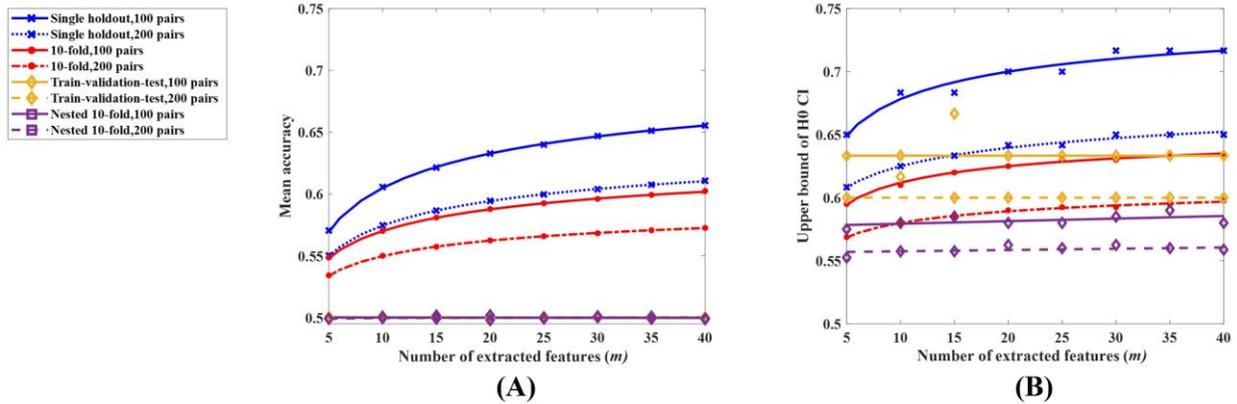

Figure 6. Effect of dimension of the feature space ($m$) on (A) mean $H_0$ accuracy and (B) upper bound of $H_0$ CI (one-sided, $\alpha = 0.05$).

Finally, the statistical powers of the different cross-validation methods were compared. To that end, the minimum required sample size for a study to detect a statistically significant effect ($\alpha = 0.05$) with typical statistical power ($1-\beta = 0.8$) was determined for each experimental condition. Figure 7 shows the results, where the markers are the outcome accuracies of the experiments, and the curves are polynomials of order 2 fitted on them. The portions of the figures that fall below the dashed black line correspond to underpowered cases. The results indicate that the train-validation-test and the single holdout methods have the lowest statistical power. For example, the train-validation-test method (Figure 7C) with 50 pairs of samples would be underpowered (it's below the dashed black line, meaning $1-\beta < 0.8$) even with two highly discriminative features (D = 1) in our 20-dimensional feature space. The situation for the single holdout method (Figure 7A) is slightly better. A different interpretation of the results would be, if we have two relatively good features with D = 0.6 in our 20-dimensional feature space, the train-validation-test needs at least 200 pairs of samples for having an 80% chance of getting a significant outcome, whereas the nested 10-fold method (Figure 7D) only needs 100 pairs of samples. Finally, the 10-fold cross-validation (Figure 7B) has slightly higher statistical power compared to the nested 10-fold method. The most likely reason for this observation would be the fact that nested *k*-fold uses fewer samples during the training and feature selection (in our case, 90%), whereas 10-fold cross-validation takes advantage of all the data.

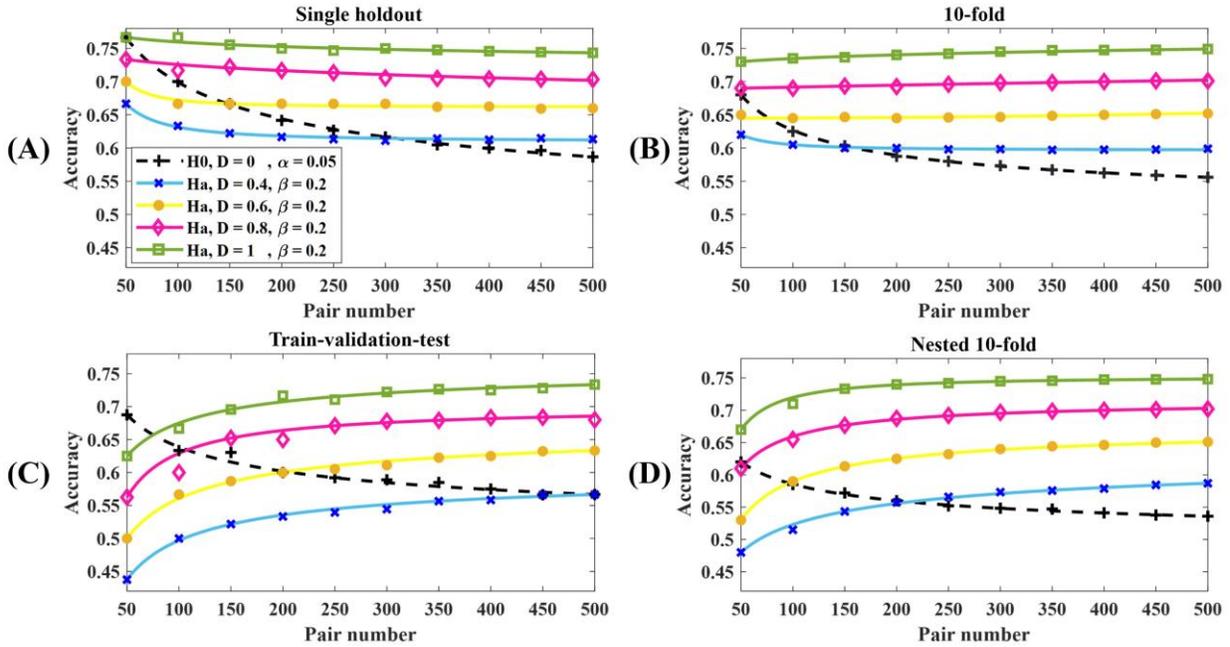

Figure 7. Comparison of the statistical power of the four cross-validation methods given varying simulated effect sizes and sample sizes. The portions of the figures that fall below the dashed black line correspond to underpowered cases: (A) single holdout, (B) 10-fold, (C) train-validation-test, (D) nested 10-fold.

**Experiment 2: Statistical confidence of the model**

Experiment 2 addressed research question Q2 ("How is the statistical confidence in an ML model affected by the employed cross-validation method?") by investigating the effect of different parameters on the statistical confidence of the final trained model. First, the effects of different cross-validation methods and sample sizes were investigated. Values of $m = 20$ (20 extracted features) and D = 0.8 were used. The generated feature set was then passed through feature selections with different cross-validations, and the two most discriminative features ($l = 2$) were determined. Each simulation was repeated 5000 times, and then the probability of finding the correct features was computed.

Figure 8(A) shows the probability of at least one of the selected features being correct ($C_{2,1}$), and 8(B) represents the probability of both selected features being correct ($C_{2,2}$). Based on these figures, models generated based on single holdout and nested 10-fold cross-validations had the lowest and highest statistical confidence, respectively. For example, the probability of both features being correct with 100 pairs is only about 20% with the single holdout (i.e., there is an 80% chance that the final model is based on at least one incorrect feature). However, this number increases to about 80% in nested 10-fold cross-validation. Additionally, comparing Figures 8(A) and 8(B) shows that the gap between the two methods increases as we aim for a correct multi-dimensional model (here the dimension of two). The nested 10-fold method has a slightly better performance compared to the 10-fold cross-validation method. This could be attributed to the voting mechanism described in the Cross-validation methods section for aggregating the outcomes of feature selection from different outer folds. Finally, the plots exhibit a saturation phenomenon, indicating there could be a "sweet spot" in terms of the recommended sample size required for getting a robust outcome.

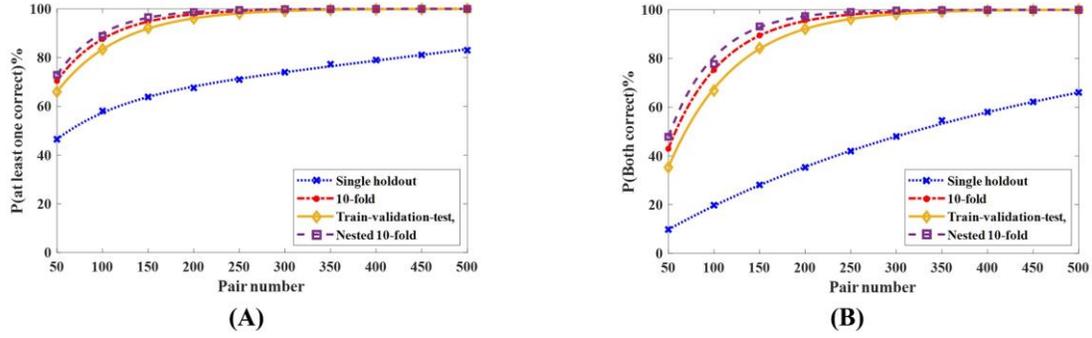

Figure 8. Effect of different cross-validation methods on the statistical confidence of the trained model: (A) $C_{2,1}$, (B) $C_{2,2}$

Next, the interaction effect between the dimensionality of the feature space ($m$) and cross-validation methods on the statistical confidence of the model was investigated. For this analysis, the value of D = 0.8 was used, and each simulation was repeated 5000 times. Figure 9(A) shows the probability of at least one of the selected features being correct ($C_{2,1}$), and Figure 9 (B) represents the probability of both selected features being correct ($C_{2,2}$). Based on these figures, the dimensionality of the feature space ($m$) and statistical confidence of the model are inversely related to each other. However, different cross-validation methods exhibit dissimilar behaviors. Specifically, whereas the single holdout is very sensitive to the dimensionality of the feature space, nested 10-fold cross-validation is very robust. Finally, there is an interaction effect between the sample size and the dimensionality of the feature space on the statistical confidence of the model. For example, if we have 100 pairs of samples and increase $m$ from 5 to 40, $C_{2,2}$ drops from 91% to 73% in nested 10-fold. However, if we increase the sample size to 200 pairs, we will only see a decrease of 4% in statistical confidence. This result indicates that, for a feature with a target effect size (D), there could be a "sweet spot" in terms of the recommended sample size required for statistical confidence to become robust to an increase in the dimensionality of the feature space.

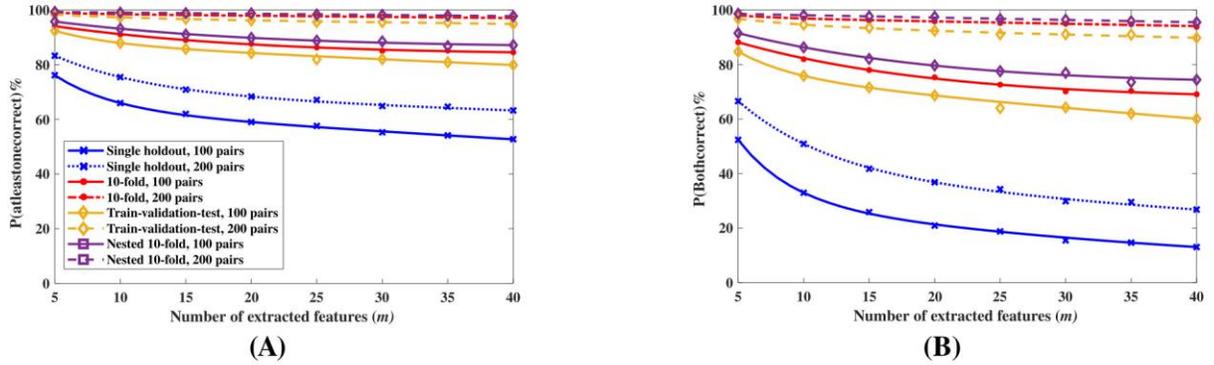

Figure 9. Effect of dimensionality of the feature space ($m$) on statistical confidence of the trained model: (A) $C_{2,1}$, (B) $C_{2,2}$

**Experiment 3: Statistical characteristics of nested $k$-fold cross-validation**

Experiments 1 and 2 showed that nested $k$-fold cross-validation offers the best statistical power and statistical confidence of the model among the investigated methods. Experiment 3 addressed research questions Q3 and Q4 by quantifying the statistical properties of nested 10-fold cross-validation and providing models that would enable future studies to estimate the required and recommended sample sizes in the design phase of the study. Given the computational cost of running nested $k$-fold and the number of required analyses for this experiment, each set of parameters was simulated 2000 times.

Based on Figure 6(B), the 95th percentile of $H_0$ for nested $k$-fold cross-validation was found to be relatively independent of the dimensionality of the feature space ($m$). We formally tested to see if the number of extracted features ($m$) and the number of selected features ($l$) had significant effects on the $H_0$ CI. To that end, a two-way analysis of variance (ANOVA) in a 4×3 design was used with independent variables $m \in [10, 20, 30, 40]$ and $l \in [2, 3, 4]$. The dependent variable was the upper bound of $H_0$ CI (one-sided, $\alpha = 0.05$). Table 2 presents the result, which did not show statistically significant effects of $m$ and $l$ on the upper bound of $H_0$ CI, confirming the robustness of the upper bound of $H_0$ CI in nested $k$-fold cross-validation to $m$ and $l$.

Table 2. Results of two-way 4×3 analysis of variance on the upper bound of the $H_0$ confidence interval.

| Predictor | Sum of squares | df | Mean square | F | p |
|---|---|---|---|---|---|
| Main effect: Feature space dimensionality ($m \in [10, 20, 30, 40]$) | 0.0017 | 3 | 0.0006 | 0.51 | 0.67 |
| Main effect: Number of selected features ($l \in [2, 3, 4]$) | 0.0042 | 2 | 0.0021 | 1.87 | 0.15 |
| Interaction effect | 0.0093 | 6 | 0.0016 | 1.37 | 0.22 |

Next, the minimum required number of pairs ($n_r$) for a study using nested 10-fold cross-validation to detect a statistically significant effect ($\alpha = 0.05$) with typical power ($1-\beta = 0.8$) was computed for $D \in [0.4, 0.5, 0.6, 0.7, 0.8, 0.9, 1], m \in [10, 20, 30, 40], l \in [2, 3, 4]$. Figure 10 shows the results, which indicate that $n$ is inversely related to D and $l$, and directly related to $m$, with D being the most influential factor. This means the function for estimating $n_r$ must integrate all three predictors.

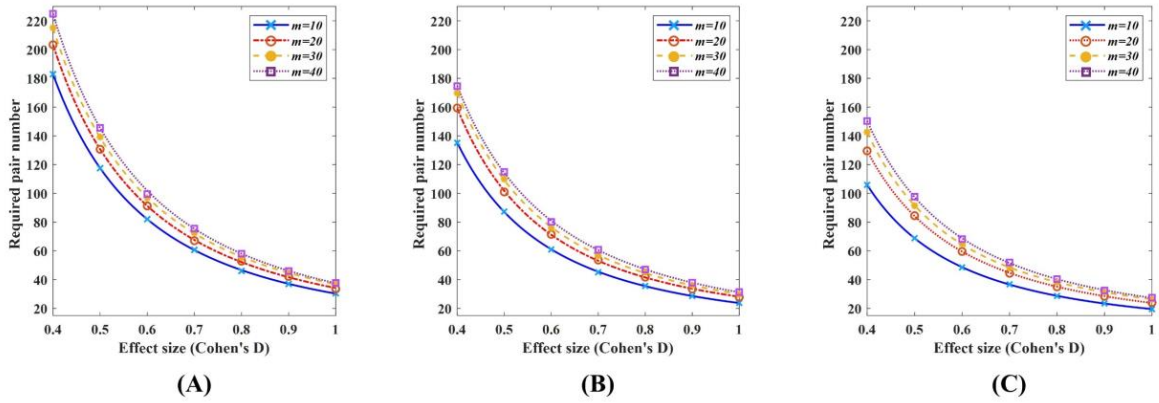

Figure 10. Effect of the size of the feature space ($m$), the number of discriminative features ($l$), and effect size (D) on the required number of pairs ($n_r$) for nested 10-fold cross-validation: (A) $l = 2$, (B) $l = 3$, (C) $l = 4$.

To derive a closed-form equation for estimating $n_r$, for each pair of $m$ and $l$, a power curve with the general form of $ax^b + c$ was used to estimate the value of $n_r$ from D. This resulted in 12 different curves. Then, the parameters of those curves ($a$, $b$, and $c$) were estimated from $m$ and $l$ based on three separate flat planes (one per parameter). Let $D_0$, $m_0$, and $l_0$ denote the target parameters of a study for which we want to estimate the minimum required number of pairs ($\alpha = 0.05$, $1-\beta = 0.8$). Equations 4–7 show the process:

$a = 39.37 - 6.718 l_0 + 0.263 m_0$ (4)
$b = -1.985 - 0.023 l_0 + 0.001 m_0$ (5)
$c = -0.886 + 1.507 l_0 - 0.015 m_0$ (6)
$n_r = a D_0^{\,b} + c$ (7)

The goodness of fit was evaluated using root-mean-square error (RMSE) and mean percent magnitude error (MPE). MPE was defined as the magnitude of the error divided by the true value expressed in percentage averaged over all testing instances . Using Equations 4–7 with the training samples (resubstitution error rate) resulted in the RMSE of 3.35 and MPE of 2.98%. That is the expected magnitude error of using equations 4–7 for estimating the number of required pairs ($n_r$) was less than 3% on the training set.

An independent test set was generated for the following sets of parameters: $D \in [0.55, 0.75, 0.95, 1.2, 1.4], m \in [20, 30, 40]$, and $l \in [2, 3, 4, 5, 6]$. Our preliminary analysis showed the value of MPE over all test samples was 11.6%, which is substantially higher than the training MPE. Therefore, further investigation was carried out to find the source of the difference. Our investigation showed correlation between MPE and the number of extracted features (i.e., $m$) was non-significant ($p = 0.13$). Similarly, correlation between MPE and discriminatory power of the features (i.e., D) was non-significant ($p = 0.63$). In contrast, MPE was highly correlated with the number of selected features (R = 0.82, p < 0.0001). Our further investigation showed that the value of MPE was quite different among the cases where $l \in [2, 3, 4]$ and $l \in [5, 6]$. Going back to Experiment 3, Equations 4–7 were trained based on data points with $l \in [2, 3, 4]$. Therefore, the test samples with $l \in [2, 3, 4]$ correspond to interpolation, whereas the test samples with $l \in [5, 6]$ correspond to extrapolation cases. The interpolation cases (45 data points) had an MPE of 3.5%, which is quite close to the MPE over the training samples, and also had a non-significant correlation coefficient ($p = 0.19$). However, the MPE for extrapolation cases (30 data points) was 23.9%, with a high correlation ($r = 0.73$, $p < 0.00001$). In summary, Equations 4–7 are very reliable for estimating the required sample size for $l \in [2, 3, 4]$, but the results become less reliable as we start to use them for extrapolating into models with higher dimensions ($l \geq 5$).

The statistical confidence of the final model ($C_{2,2}$, that is, the probability of both selected features being correct) using nested 10-fold cross-validation was computed for different numbers of pairs and $D \in [0.4, 0.5, 0.6, 0.7, 0.8, 0.9, 1], m \in [10, 20, 30, 40], l = 2$. The results are presented in appendix Tables A1 and A2 with selected cases ($D \in [0.4, 0.6, 0.8, 1], m = 10, l = 2$) plotted in Figure 11.

Finally, the recommended sample size in terms of the number of pairs (*n*) for achieving a target confidence for the final model ($C_{2,2}$) can be obtained from Tables A1 and A2 in conjunction with linear interpolations. For example, if we have two relatively good features with D = 0.6 in our 10-dimensional feature space, and we want to reach the confidence of 95%, we need to have about 215 pairs of samples. However, to achieve the same confidence with 40 extracted features, 342 pairs of samples need to be used.

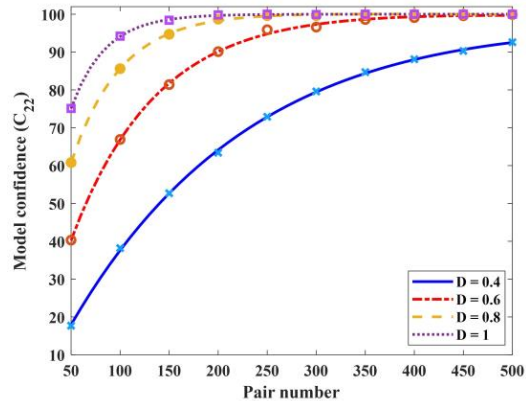

Figure 11. Effect of D and *n* on the confidence of the model generated using nested 10-fold cross-validation when *m* = 10.

**DISCUSSION**

This study was primarily motivated by the observation that many ML-based studies in speech, language, and hearing sciences have adopted cross-validation methods with a single test set (i.e., single holdout or train-validation-test cross-validation approaches). Our initial investigation using real clinical data showed a profound impact of employed cross-validation on the estimated accuracy and its dispersion, and the stability of the feature selection outcome. Therefore, our main hypothesis was that the statistical properties of the final trained model would be significantly affected by which cross-validation method was used, especially at smaller sample sizes (less than a few hundred) that are the most likely scenarios in these fields.

Three experiments were conducted to quantify the impact of different cross-validation methods on the statistical properties of the final trained models and to provide methods that would enable future studies to estimate the required and recommended sample sizes during the design phase based on the most robust cross-validation method that was evaluated in the current investigation (nested *k*-fold cross-validation). Table 3 presents a summary comparing the statistical properties of the four cross-validation approaches studied. We would like to emphasize that the three factors of bias ($H_0$ mean accuracy), statistical power, and statistical confidence need to be considered simultaneously. For example, while 10-fold cross-validation has the highest power, its estimated accuracy (if the ML processing pipeline contains a model selection component) is biased, and hence should not be used. Conversely, despite the fact that train-validation-test method gives an unbiased estimate of the performance, the method has very low statistical power, meaning the distribution of $H_0$ and $H_a$ have a lot of overlap. Therefore, there is a significant probability for random features to produce high classification accuracies with train-validation-test method (please refer to Figure 5(A) and Figure 7). It is hoped that the results of this work will provide strong motivation for researchers to move away from single holdout and train-validation-test cross-validation approaches and instead consider using the unbiased and more robust nested *k*-fold cross-validation method in future studies.

Table 3. A summary comparing the four cross-validation methods studied. Desirable model characteristics are marked with bold face letters.

| Method | $H_0$ mean accuracy | $H_0$ std accuracy | Robustness to large feature space | Statistical power | Statistical confidence | Required sample size | Recommended sample size | Computational time |
|---|---|---|---|---|---|---|---|---|
| Single holdout | Highest bias | Low | Lowest | Low | Lowest | High | Highest | **Lowest** |
| $k$-fold | High bias | **Lowest** | High | **Highest** | High | **Lowest** | Low | Low |
| Train-validation-test | **Unbiased** | Highest | Moderate | Lowest | Moderate | Highest | Moderate | Low |
| Nested $k$-fold | **Unbiased** | Moderate | **Highest** | High | **Highest** | Low | **Lowest** | Highest |

**Q1: How is the statistical power of an ML model affected by the employed cross-validation method?**

Experiment 1 answered this question by comparing the distribution of the null hypothesis $H_0$ (i.e., when the features were irrelevant) and alternative hypothesis $H_a$ (when some of the features were discriminative) for four different cross-validation methods. Results showed that the widely used single holdout method has significant bias (Figure 5(A)) and therefore can significantly overestimate the true performance; at the same time, single holdout cross-validation leads to an $H_0$ with a very heavy tail (Figure 5(B)). This means that, even if the features are truly random (irrelevant features), there is a good chance for the single holdout method to produce a high level of accuracy. On the other hand, not only does nested $k$-fold cross-validation give an unbiased estimate of the performance (Figure 5(A)), but also its $H_0$ distribution has a thinner higher tail (Figure 5(B)). This means that the probability of getting a very high accuracy from irrelevant features from nested $k$-fold cross-validation is much lower than the single holdout cross-validation.

Another advantage of nested $k$-fold cross-validation is higher robustness to the dimensionality of the feature space. This means that the probability of getting high accuracy from irrelevant features increases with an increase in the number of extracted features if the single holdout method is used; in contrast, increasing the number of features does not artificially increase accuracy if nested $k$-fold cross-validation is used (Figure 6(B)). Considering that most ML-based studies are exploratory in nature, they often start with a relatively high number of features, and, therefore, using a method that is less affected by the number of extracted features is quite desirable.

The results of Experiment 1 also demonstrated that robust statistical results could be obtained with significantly smaller required sample sizes when using nested $k$-fold cross-validation as compared to the single holdout ot train-validation-test methods (Figure 7). Considering that data collection is usually the most costly part of a clinical study in terms of time and expense, a significant reduction in the required sample size could increase the likelihood of carrying out a study that may have previously been deemed impractical or too costly. This reduction of the required sample size of course comes at the expense of increased computational complexity of the nested $k$-fold cross-validation method. However, the availability of fast computers and high-performance computing resources reduces this concern, and the trade-off is quite favorable.

**Q2: How is the statistical confidence in an ML model affected by the employed cross-validation method?**

Statistical confidence of the model was defined as the probability of correct features being selected and, hence, contributing to the classification outcome. Experiment 2 answered research question Q2 by quantifying the effect of the four cross-validation methods on the statistical confidence of the model. Based on Figure 8, the statistical confidence of the model based on nested $k$-fold cross-validation could be as much as 4 times higher than the confidence associated with the model based on single holdout cross-validation. The other important finding was the dissimilar behavior of statistical confidence of the model when the number of extracted features increases. Based on Figure 9(B) not only does the nested $k$-fold cross-validation lead to significantly higher confidence in the model, but it is also more robust than the other methods to the dimensionality of the feature space. Therefore, whereas we need to be extra cautious and extract "as few" features as possible when employing single holdout cross-validation, we could relax this requirement and extract more features with nested $k$-fold cross-validation. In practice, this relaxation could significantly increase the chance of finding more powerful and discriminative features. These important characteristics should provide further motivation for shifting from the single holdout method to nested $k$-fold cross-validation.

Based on the results of Experiments 1 and 2, the nested $k$-fold cross-validation method exhibited the best performance, followed by the 10-fold cross-validation method. Train-validation-test cross-validation, which is a popular method in the machine learning community due to its unbiased estimate of accuracy (see Figure 5(A)) and the simplicity of its implementation, had the lowest statistical power (Figure 7) and needed even more samples compared to the single holdout method. However, the statistical confidence of the model generated based on the train-validation-test was much higher than the single holdout method (Figure 8). This higher confidence is attributed to our implementation of train-validation-test cross-validation. Specifically, the train-validation part was split using 10-fold cross-validation in our implementation, and, based on Figure 8, employing 10-fold cross-validation leads to models with high statistical confidence. Then, the inferior statistical confidence of the model in the train-validation-test compared to the 10-fold cross-validation is attributed to the fact that the train-validation-test method only uses a portion of data for feature selection (85% in our implementation). However, if the train-validation part was split using a single holdout method (which in fact is very common in practice), the statistical confidence of the model based on the train-validation-test would have been lower than the statistical confidence of the single holdout method. In summary, the findings of this study advise against any application of single holdout and train-validation-test altogether.

**Q3: What is the minimum required sample size to get a statistically significant outcome from an ML model with conventional power requirements (5% significance, 80% power [α=0.05, 1-β=0.8])?**

Experiment 3 was devoted to answering research questions Q3 and Q4 for a model employing nested 10-fold cross-validation, which was the method with the best statistical properties. This model was described in Equations 4–7. The discriminative power of selected features (D) was the most influential factor in determining the minimum required sample size, followed by the dimensionality of the model ($l$), and then the number of extracted features ($m$). Additionally, the minimum required sample size in terms of the number of pairs ($n_r$) decreases as the discriminative power of selected features (D) or dimensionality of the model ($l$) increases. In contrast, the minimum required sample size increases as the number of extracted features ($m$) increases.

This last observation may seem to contradict the robustness of the distribution of the null hypothesis regarding the number of extracted features (Figure 6(B) and Table 2) and warrants further explanation. The minimum required sample size depends on the distributions of both null and alternative hypotheses. The distribution of the null hypothesis of nested $k$-fold cross-validation is relatively invariant to the number of extracted features (similar value of $\alpha$ for a fixed threshold). However, as more features get extracted (a larger $m$), the distribution of the alternative hypothesis changes and gets more overlapped with that of the null hypothesis ($\beta$ increases). Therefore, the required sample size needs to be increased to reduce the value of $\beta$ and attain similar statistical power.

Lastly, an example is provided to show how Equations 4–7 can be used to estimate the required sample size during the experimental design. Let's consider the example presented in "COMPARING CROSS-VALIDATION METHODS IN A REAL-WORLD CLINICAL EXAMPLE" section, where we know we want to extract 48 features prior to the execution of the study. Prior studies have reported the effect sizes in the range of 0.43–0.88 for such a problem (Van Stan et al., 2020). Plugging $D_0 = 0.66$ (as an average expected effect size), $m_0 = 48$, and $l_0 = 2$ into Equations 4–7, we get $n_r = 89.3$, meaning that we need to recruit a minimum of 90 pairs of participants to have the adequate power to run the analysis.

**Q4: What is the recommended sample size for achieving a target statistical confidence of the final ML model?**

The second part of Experiment 3 provided an answer to Q4 for nested 10-fold cross-validation. While we were not able to derive a closed-form formula for this case, different tables and MATLAB code are provided that can be used for linear interpolation and estimation of the recommended sample size for target values of D and $m$. It is noteworthy that using Tables A1 and A2 for estimating the recommended number of pairs ($n$) will result in a larger value compared to Equations 4–7. For example, we only need 89 pairs for D = 0.6, $m$ = 20, and $l$ = 2 to obtain a statistically significant effect ($\alpha$ = 0.05, 1−$\beta$ = 0.8). However, 89 pairs with the same parameters will provide 50% confidence in the model. To achieve model confidence of 80% and 95%, respectively, the sample size needs to be increased to 169 and 247 pairs. This wide gap is because Equations 4–7 estimate the required $n$ for just having the statistical power to detect a significant effect. Whereas, having a model with high statistical confidence (a large $C_{2,2}$) is a more restrictive requirement than getting a statistically significant outcome and, therefore, requires more samples. The number of pairs necessary for reaching a certain confidence in the model increases with a decrease in the discriminative power of features (D) and with an increase in the number of extracted features ($m$). As a final note, considering the negative effect of the high number of extracted features on the statistical power and statistical confidence of the model, keeping the number of extracted features relatively low is preferable.

ML solutions often have different optimization processes. The focus of this study was to quantify the statistical characteristics of different cross-validation methods in the presence of feature selection. However, other processes including hyperparameter optimization and optimization of the architecture (e.g., in deep learning) are also important and widely used in the ML community. While the specific findings (e.g., the method for estimating the required sample size) of this study are limited to cases with feature selection, the general findings can be applied to other optimization scenarios. For example, single holdout or train-validation-test cross-validation methods should not be used for hyperparameter optimization and optimization of the architecture of deep networks. Instead, the nested *k*-fold cross-validation method should be employed. This argument is especially important for research with small sample sizes (less than a few hundred).

**Computing appropriate dimensionality of the feature space**

The presented power analysis method (Equations 4–7) was derived to assist researchers with determining the minimum sample size required during experimental design; however, the method can also be used for existing datasets. Specifically, during experimental design, researchers know the features that they want to use so the dimensionality of the feature space can be estimated, and their goal is to determine the required sample size. However, the situation reverses with existing datasets. That is, the sample size is already known, and the goal is to determine the appropriate dimensionality of the feature space. The presented method can also be used to estimate the maximum allowable dimensionality of the feature space. If the number of extracted features were higher than this estimated number, some features would need to be eliminated such that the dimensionality of the feature space gets below the maximum allowable dimension. We would like to emphasize that this elimination should be carried out independently from the labels of the dataset; otherwise, the elimination procedure will most likely overestimate the performance of the model and will reduce the generalizability of the findings. Some examples of appropriate elimination would be using the literature (generated from different datasets) for removing inferior features, computing correlation between features of the dataset and only keeping one among highly correlated features, or applying an unsupervised dimensionality reduction without any optimization based on the labels of the dataset.

Lastly, an example is provided to show how Equations 4–7 can be used to estimate appropriate dimensionality of the feature space for an existing dataset. Let's consider the example presented in "COMPARING CROSS-VALIDATION METHODS IN A REAL-WORLD CLINICAL EXAMPLE" section, where our existing dataset has 136 pairs of participants. Plugging $D_0 = 0.66$ (as an average expected effect size), $m_0 = 135$, and $l_0 = 2$ into Equations 4–7 we get $n_r = 135.2 \approx 136$, meaning that with 136 pairs of participants the maximum allowable dimensionality of the feature space would be 135.

**Limitations**

Despite the importance of our current findings, several limitations should be mentioned. Our findings were based on Monte Carlo simulations, and we had to make certain assumptions to keep the complexity of our simulations manageable. Those assumptions are responsible for most of the limitations of our findings. The first assumption was that ML models were being used for a binary classification problem. Although the findings would still be valid for a multi-class classification problem (e.g., that the single holdout method is the least reliable and nested *k*-fold cross-validation is the most reliable method), the results of Experiment 3 could not be applied directly to other classification problems and need to be simulated separately. The second assumption was that of a balanced dataset (equal number of samples in both classes), which in practice may not be achievable or desirable. The third assumption was that all of the discriminative features each had equal discriminative power represented by the same value of D. The fourth assumption was that all discriminative features had Gaussian distributions, and, hence, a linear classifier (logistic regression) was able to find the optimum decision boundary. The fifth assumption was that none of the extracted features were correlated, and the remaining ($m-l$) features were irrelevant to the classification problem. While it is desirable to extract only features that add novel information to the model and are uncorrelated, in practice, this is hardly ever achieved. Finally, it is important to note that the estimation of the minimum sample size of conventional statistical tests (e.g., t-test, ANOVA, correlation, etc.) shares many of these assumptions and limitations, and, in that regard, the methodology presented in Experiment 3 extends the application of widely used power analysis to include ML models.

**Generalizability of results to unbalanced datasets and other classification frameworks**

As described in the Limitations section, the equations for the estimation of the required sample size were derived based on simulated balanced datasets with underlying Gaussian distributions and with equal discriminative power for selected features. This section includes modifications that would allow applications of power analysis to unbalanced datasets and features with unequal discriminative powers. We will also present preliminary results on a direction that could extend the application of the method to non-linear feature spaces.

First, unbalanced datasets were investigated. Let $\gamma_{Db}$ denote the ratio between the number of samples in the positive ($n_{positive}$) and negative classes ($n_{negative}$). The following sets of parameters were used: $m = 20, l = 2, D \in [0.6, 0.8, 1]$, and $\gamma_{Db} \in [1.2, 1.4, 1.6, 1.8, 2]$. Two different scenarios were tested. In the first scenario, Equations 4–7 were used for estimating the required sample size of the smaller class. The error was always negative, with MPE equal to 44.2%. This indicates that using Equations 4–7 for estimating the required sample size of the smaller class leads to a significant overestimation. In the second scenario, Equations 4–7 were used for estimating an adjusted required sample size, where the adjusted required sample size was defined as the average of the sample sizes of the positive and negative classes ($\frac{n_{negative}+n_{positive}}{2}$). The error was reduced, with MPE equal to 12.4%. Further investigation indicated that, as the value of D increases, estimation of the adjusted required sample size becomes more robust to higher imbalance ratios ($\gamma_{Db}$). For example, if D = 0.6 and $\gamma_{Db} = 2$, MPE for the adjusted required sample size was 70.6%; however, for D = 0.8 and $\gamma_{Db} = 2$, MPE was reduced to 16%. Therefore, Equations 4–7 are relatively robust to moderately imbalanced datasets when employing the adjusted required sample size and if features are highly discriminative (D ≥ 0.8). The figures for this analysis are provided in Figure A1 in the Appendix. In summary, the "Compute_RequiredSampleSize.m" code provided along this manuscript can be used for unbalanced datasets, however, the estimated required sample size would be the average of the number of negative and positive samples. Estimation of the sample size of each class from the estimated average required sample size and $\gamma_{Db}$ should be trivial.

Second, a limited number of cases with features with unequal effect sizes (D) were simulated. To that end, a series of 20-dimensional feature spaces ($m = 20$) containing two discriminative features ($l = 2$) were generated. The discriminatory power of the first feature was taken from the set $D \in [0.6, 0.8, 1]$, and the discriminatory power of the second feature was set equal to $\gamma_D \times D$. The parameter $\gamma_D$ captures the unequal discriminatory power between the two features, and $\gamma_D \in [1.2, 1.4, 1.6, 1.8, 2]$. The average discriminatory power of the two features ($D_0 = D\frac{(\gamma_D+1)}{2}$) was used in Equation 7. The value of MPE was 7.9%. Our further investigation indicated that, as the value of D increases, estimation of the required sample size becomes more robust to higher imbalance ratios ($\gamma_D$). For example, if D = 0.6 and $\gamma_D = 2$, MPE was 23.1%. However, for D = 0.8 and $\gamma_D = 2$, MPE was reduced to 16.7%. Therefore, Equations 4–7 are relatively robust to features with different discriminatory powers if both features are discriminative enough (D ≥ 0.8). The figures for this analysis are provided in Figure A2 in the Appendix. In summary, the "Compute_RequiredSampleSize.m" code provided along this manuscript can be used for features with different unequal discriminative powers, however, the average of their discriminative powers needs to be used.

Finally, the possibility of applying the method to non-linear feature spaces and other classification algorithms was investigated. The findings of this study were based on simulated datasets with underlying Gaussian distributions. In that sense, quantification of their discriminative power using Cohen's D and the application of logistic regression were optimum selections. We would like to emphasize another rationale for parameterization of power analysis in terms of Cohen's D, and then we will present a possible direction for the generalization of power analysis to features with non-Gaussian distributions and non-linear classifiers. Researchers often use effect sizes reported in the literature for estimating the minimum required sample size during the design of a study, and, currently, the application of Cohen's D is ubiquitous in clinical sciences (Fritz et al., 2012). Therefore, the selection of Cohen's D was driven by this important practical consideration.

We will now present a possible modification that could extend the application of the presented power analysis to more general cases. However, an in-depth treatment of this topic is out of the scope of this paper and needs to be pursued in an independent study. Our proposed solution is based on computing an "equivalent Cohen's D" for such general cases. The "equivalent Cohen's D" for a non-linear feature space is the value of D that, if used to generate Gaussian features (please refer to Equations 1 and 2), produces similar classification accuracy with logistic regression as the problem under investigation. We will demonstrate this approach with bull's eye datasets with five different noise levels of [2, 2.5, 3, 3.5, 4]. Different noise levels model different amounts of overlap between the two classes. Specifically, for a noise level of 0, the dataset will be two perfect co-centric circles with a between-class distance of 1 unit. The noise level of 2 corresponds to a uniform disturbance in the range of -1 to 1 (i.e., equal to the true between-class distance). The noise level of 4 corresponds to a uniform disturbance with a range that is two times larger than the true between-class distance. Figure A3 in the Appendix shows examples of this dataset with low and high noise levels. A support vector machine (SVM) with a polynomial kernel was adopted for learning the decision boundary between the two classes. For each noise level, the two dimensions of the bull's eye dataset were augmented with 18 random features to create a 20-dimensional feature space ($m = 20$). Nested 10-fold cross-validation with polynomial SVM was used to find the two most discriminative features. Following the methodology presented in the "Statistical power and the required sample size" section (please refer to Figure 4), the true required sample size for rejecting the null hypothesis was determined for each noise level. The equivalent Cohen's D of each noise level was estimated from a look-up table that we generated from simulations of Gaussian features with a logistic regression classifier. The computed equivalent Cohen's D was then used to estimate the required sample size from Equations 4–7. The results showed an MPE of 8.6%, confirming the potential of this approach.

**Directions for future studies**

The study presented here can be extended in several directions. The first direction would be a rigorous and in-depth treatment of extending the power analysis to non-linear feature spaces. Updating the power analysis method to better account for unbalanced datasets and unequal discriminative power for selected features are other directions. The current study showed that statistical properties of nested $k$-fold cross-validation are very desirable. However, only the value of $k = 10$ was implemented. While the general findings of the study should hold for other values of $k$, we expect an interaction effect among the power of the method, the sample size, and the value of $k$. A future study may suggest the appropriate value of $k$ for a given problem such that the statistical power of the model is maximized. The current study used Cohen's D for Gaussian features and proposed equivalent Cohen's D for general feature spaces. However, investigation of the most appropriate way to quantify and report the effect size in ML would allow practitioners and researchers to use prior studies during experimental design and is a question for future research. A more theoretical treatment of power analysis and all of the above-mentioned directions could significantly help scientists and practitioners with the appropriate and responsible usage of ML. Finally, evaluating the effect of different cross-validation methods on the performance of ML using Bayes Error is another direction for future studies.

**CONCLUSION**

Monte Carlo simulations were adopted in this study to quantify the statistical characteristics of different cross-validation methods. Our analyses showed significantly different statistical characteristics among different cross-validation methods. The model based on single hold-out had very low statistical power (meaning it required many more samples to get a statistically significant outcome), very low statistical confidence (meaning it was very likely for incorrect features to be selected), and very high bias (meaning it was very likely to overestimate the performance of the model). The statistical characteristics of the popular method of train-validation-test were also very poor. Conversely, the model based on nested 10-fold cross-validation resulted in the highest statistical power and statistical confidence. Also, the estimated performance was unbiased. Our numerical analyses showed that the required sample size with a single holdout could be 50% higher than what would be needed if nested $k$-fold cross-validation were used. Also, statistical confidence in the model based on nested $k$-fold cross-validation was as much as four times higher than the statistical confidence in the single holdout-based model. Finally, computational models along with MATLAB code were provided to assist in power analysis for designing future ML studies to determine the required and recommended sample sizes when employing nested 10-fold cross-validation. Based on these findings, researchers are highly encouraged to avoid using the single holdout and train-validation-test methods and use the unbiased and more robust method of nested $k$-fold cross-validation. As a final note, if the ML processing pipeline does not include any model selection component (e.g., feature selection, hyperparameter optimization, optimization of the architecture of a deep or shallow neural network, selecting between different classification algorithms, etc.), $k$-fold cross-validation (no nesting) may be used due to its lower computational complexity relative to nested $k$-fold cross-validation.

**ACKNOWLEDGMENTS**

Research reported in this publication was supported by the National Institute On Deafness And Other Communication Disorders of the National Institutes of Health under Award Numbers (T32 DC013017, P50 DC015446, K99 DC021235). The content is solely the responsibility of the authors and does not necessarily represent the official views of the National Institutes of Health. The authors would like to acknowledge the MGH high-performance computing resource that made the execution of our intensive simulations possible.

DISCLOSURE

Drs. Robert Hillman and Daryush Mehta have a financial interest in InnoVoyce LLC, a company focused on developing and commercializing technologies for the prevention, diagnosis, and treatment of voice-related disorders. Dr. Hillman's and Dr. Mehta's interests were reviewed and are managed by Massachusetts General Hospital and Mass General Brigham in accordance with their conflict-of-interest policies.

DATA AVAILABILITY STATEMENT

Mass General Brigham and Mass General are not allowed to give access to data without the Principal Investigator (PI) for the human studies protocol first submitting a protocol amendment to request permission to share the data with a specific collaborator on a case-by-case basis. This policy is based on strict rules dealing with the protection of patient data and information. Anyone wishing to request access to the data used for comparing cross-validation methods in a real-world clinical example section must contact Ms. Sarah DeRosa (sederosa@partners.org), Program Coordinator for Research and Clinical Speech-Language Pathology, Center for Laryngeal Surgery and Voice Rehabilitation, Massachusetts General Hospital.

The data used for Experiments 1–3 was generated using Monte Carlo simulations. Several MATLAB functions associated with this manuscript are available. The latest version of the code can be retrieved from https://github.com/GhasemzadehHamzeh/ML_PowerAnalysis.

The "Compute_RequiredSampleSize.m" code takes the number of extracted features ($m\_0$), the effect size of the best features ($D\_0$), and the number of selected features ($l\_0$) as the input parameters and computes the required number of pairs for having the statistical power for finding a statistically significant model ($\alpha = 0.05$, $1-\beta = 0.8$). This code can also be used for unbalanced datasets; however, the estimated required sample size ($n_r$) would be the average of the number of negative and positive samples. The sample size of each class can then be computed using the estimated average required sample size and $\gamma_{Db}$. Specifically, the required sample size of the smaller class will be $n_r \times \frac{2}{1+\gamma_{Db}}$ and the required sample size of the larger class will be $n_r \times \frac{2 \times \gamma_{Db}}{1+\gamma_{Db}}$. Similarly, this code can also be used for features with different effect sizes; the average of the effect sizes needs to be input as $D\_0$.

The "Compute_RecommendedSampleSize.m" code takes the number of extracted features ($m\_0$), the effect size of the best features ($D\_0$), and the target value of $C_{2,2}$ ($CI\_0$) as the input parameters and computes the recommended number of pairs for achieving the target value of $C_{2,2}$.

The "Compute_NestedModelConfidence.m" code takes the number of extracted features ($m\_0$), the effect size of the best features ($D\_0$), and the number of pairs in a dataset ($n\_0$) as the input parameters and computes the value of $C_{2,2}$.

The "Feature_Selection" toolbox provides an implementation of forward feature selection with four investigated cross-validation approaches. Please, refer to the "Main_Test.m" file for examples of the usage of the code.

REFERENCES

Alharbi, S., Hasan, M., Simons, A. J. H., Brumfitt, S., & Green, P. (2020). Sequence labeling to detect stuttering events in read speech. *Computer Speech & Language*, *62*, 101052.

Arjmandi, M. K., & Pooyan, M. (2012). An optimum algorithm in pathological voice quality assessment using wavelet-packet-based features, linear discriminant analysis and support vector machine. *Biomedical Signal Processing and Control*, *7*(1), 3–19. https://doi.org/10.1016/j.bspc.2011.03.010

Arjmandi, M. K., Pooyan, M., Mikaili, M., Vali, M., & Moqarehzadeh, A. (2011). Identification of voice disorders using long-time features and support vector machine with different feature reduction methods. *Journal of Voice*, *25*(6), e275–e289.

Armstrong, R., Symons, M., Scott, J. G., Arnott, W. L., Copland, D. A., McMahon, K. L., & Whitehouse, A. J. O. (2018). Predicting language difficulties in middle childhood from early developmental milestones: A comparison of traditional regression and machine learning techniques. *Journal of Speech, Language, and Hearing Research*, *61*(8), 1926–1944.

Balki, I., Amirabadi, A., Levman, J., Martel, A. L., Emersic, Z., Meden, B., Garcia-Pedrero, A., Ramirez, S. C., Kong, D., Moody, A. R., & others. (2019). Sample-size determination methodologies for machine learning in medical imaging research: a systematic review. *Canadian Association of Radiologists Journal*, *70*(4), 344–353.

Bayerl, S. P., Wagner, D., Nöth, E., & Riedhammer, K. (2022). Detecting dysfluencies in stuttering therapy using wav2vec 2.0. *Interspeech*, 2868–2872.

Berisha, V., Krantsevich, C., Hahn, P. R., Hahn, S., Dasarathy, G., Turaga, P., & Liss, J. (2021). Digital medicine and the curse of dimensionality. *NPJ Digital Medicine*, *4*(1), 153.

Bhat, G. S., Shankar, N., & Panahi, I. M. S. (2020). Automated machine learning based speech classification for hearing aid applications and its real-time implementation on smartphone. *2020 42nd Annual International Conference of the IEEE Engineering in Medicine & Biology Society (EMBC)*, 956–959.

Bing, D., Ying, J., Miao, J., Lan, L., Wang, D., Zhao, L., Yin, Z., Yu, L., Guan, J., & Wang, Q. (2018). Predicting the hearing outcome in sudden sensorineural hearing loss via machine learning models. *Clinical Otolaryngology*, *43*(3), 868–874.

Cho, W. K., & Choi, S.-H. (2020). Comparison of convolutional neural network models for determination of vocal fold normality in laryngoscopic images. *Journal of Voice*.

Cohen, J. (2013). *Statistical power analysis for the behavioral sciences*. Academic press.

Cortés, J. P., Espinoza, V. M., Ghassemi, M., Mehta, D. D., Van Stan, J. H., Hillman, R. E., Guttag, J. V, & Zanartu, M. (2018). Ambulatory assessment of phonotraumatic vocal hyperfunction using glottal airflow measures estimated from neck-surface acceleration. *PLoS One*, *13*(12), e0209017.

Crippa, A., Salvatore, C., Perego, P., Forti, S., Nobile, M., Molteni, M., & Castiglioni, I. (2015). Use of machine learning to identify children with autism and their motor abnormalities. *Journal of Autism and Developmental Disorders*, *45*, 2146–2156.

Donohue, C., Khalifa, Y., Mao, S., Perera, S., Sejdić, E., & Coyle, J. L. (2021). Characterizing swallows from people with neurodegenerative diseases using high-resolution cervical auscultation signals and temporal and spatial swallow kinematic measurements. *Journal of Speech, Language, and Hearing Research*, *64*(9), 3416–3431.

Field, A., Miles, J., & Field, Z. (2012). *Discovering statistics using R*. Sage publications.

Figueroa, R. L., Zeng-Treitler, Q., Kandula, S., & Ngo, L. H. (2012). Predicting sample size required for classification performance. *BMC Medical Informatics and Decision Making*, *12*, 1–10.

Fritz, C. O., Morris, P. E., & Richler, J. J. (2012). Effect size estimates: current use, calculations, and interpretation. *Journal of Experimental Psychology: General*, *141*(1), 2.

Ghasemzadeh, H. (2019a). Calibrated steganalysis of mp3stego in multi-encoder scenario. *Information Sciences*, *480*, 438–453.

Ghasemzadeh, H. (2019b). Multi-layer architecture for efficient steganalysis of UnderMp3Cover in multiencoder scenario. *IEEE Transactions on Information Forensics and Security*, *14*(1), 186–195. https://doi.org/10.1109/TIFS.2018.2847678

Ghasemzadeh, H., & Arjmandi, M. K. (2020). Toward Optimum Quantification of Pathology-induced Noises: An Investigation of Information Missed by Human Auditory System. *IEEE/ACM Transactions on Audio, Speech, and Language Processing*, *28*, 519–528.

Ghasemzadeh, H., Deliyski, D. D., Hillman, R. E., & Mehta, D. D. (2021). Method for horizontal calibration of laser-projection transnasal fiberoptic high-speed videoendoscopy. *Applied Sciences*, *11*(2), 822. https://doi.org/10.3390/app11020822

Ghasemzadeh, H., Deliyski, D., Ford, D., Kobler, J. B., Hillman, R. E., & Mehta, D. D. (2020). Method for Vertical Calibration of Laser-Projection Transnasal Fiberoptic High-Speed Videoendoscopy. *Journal of Voice*, *34*(6), 847–861.

Ghasemzadeh, H., Doyle, P. C., & Searl, J. (2022). Image representation of the acoustic signal: An effective tool for modeling spectral and temporal dynamics of connected speech. *The Journal of the Acoustical Society of America*, *152*(1), 580–590.

Ghasemzadeh, H., & Searl, J. (2018). Modeling dynamics of connected speech in time and frequency domains with application to ALS. *11th International Conference on Voice Physiology and Biomechanics (ICVPB)*, August.

Ghasemzadeh, H., Tajik Khass, M., Khalil Arjmandi, M., & Pooyan, M. (2015). Detection of vocal disorders based on phase space parameters and Lyapunov spectrum. *Biomedical Signal Processing and Control*, *22*, 135–145. https://doi.org/10.1016/j.bspc.2015.07.002

Ghassemi, M., Van Stan, J. H., Mehta, D. D., Zañartu, M., Cheyne II, H. A., Hillman, R. E., & Guttag, J. V. (2014). Learning to detect vocal hyperfunction from ambulatory neck-surface acceleration features: Initial results for vocal fold nodules. *IEEE Transactions on Biomedical Engineering*, *61*(6), 1668–1675.

Gómez, P., Schützenberger, A., Semmler, M., & Döllinger, M. (2018). Laryngeal pressure estimation with a recurrent neural network. *IEEE Journal of Translational Engineering in Health and Medicine*, *7*, 1–11.

Hamed Mozaffari, M., & Lee, W.-S. (2019). Domain adaptation for ultrasound tongue contour extraction using transfer learning: A deep learning approach. *The Journal of the Acoustical Society of America*, *146*(5), EL431–EL437.

Hawkins, D. M. (2004). The problem of overfitting. *Journal of Chemical Information and Computer Sciences*, *44*(1), 1–12.

Huang, G. B., Mattar, M., Berg, T., & Learned-Miller, E. (2008). Labeled faces in the wild: A database for studying face recognition in unconstrained environments. *Workshop on Faces in'Real-Life'Images: Detection, Alignment, and Recognition*.

Ibarra, E. J., Parra, J. A., Alzamendi, G. A., Cortés, J. P., Espinoza, V. M., Mehta, D. D., Hillman, R. E., & Zañartu, M. (2021). Estimation of subglottal pressure, vocal fold collision pressure, and intrinsic laryngeal muscle activation from neck-surface vibration using a neural network framework and a voice production model. *Frontiers in Physiology*, 1419.

Jones, S., Carley, S., & Harrison, M. (2003). An introduction to power and sample size estimation. *Emergency Medicine Journal: EMJ*, *20*(5), 453.

Kapoor, S., & Narayanan, A. (2022). Leakage and the reproducibility crisis in ML-based science. *ArXiv Preprint ArXiv:2207.07048*.

Kist, A. M., Gómez, P., Dubrovskiy, D., Schlegel, P., Kunduk, M., Echternach, M., Patel, R., Semmler, M., Bohr, C., Dürr, S., & others. (2021). A deep learning enhanced novel software tool for laryngeal dynamics analysis. *Journal of Speech, Language, and Hearing Research*, *64*(6), 1889–1903.

Krstajic, D., Buturovic, L. J., Leahy, D. E., & Thomas, S. (2014). Cross-validation pitfalls when selecting and assessing regression and classification models. *Journal of Cheminformatics*, *6*(1), 1–15.

Le, D., Licata, K., Persad, C., & Provost, E. M. (2016). Automatic assessment of speech intelligibility for individuals with aphasia. *IEEE/ACM Transactions on Audio, Speech, and Language Processing*, *24*(11), 2187–2199.

Lenatti, M., Moreno-Sánchez, P. A., Polo, E. M., Mollura, M., Barbieri, R., & Paglialonga, A. (2022). Evaluation of machine learning algorithms and explainability techniques to detect hearing loss from a speech-in-noise screening test. *American Journal of Audiology*, *31*(3S), 961–979.

Liu, X., Faes, L., Kale, A. U., Wagner, S. K., Fu, D. J., Bruynseels, A., Mahendiran, T., Moraes, G., Shamdas, M., Kern, C., & others. (2019). A comparison of deep learning performance against health-care professionals in detecting diseases from medical imaging: a systematic review and meta-analysis. *The Lancet Digital Health*, *1*(6), e271–e297.

Lowry, R. (2014). *Concepts and applications of inferential statistics*.

Marcot, B. G., & Hanea, A. M. (2021). What is an optimal value of k in k-fold cross-validation in discrete Bayesian network analysis? *Computational Statistics*, *36*(3), 2009–2031.

Matikolaie, F. S., & Tadj, C. (2022). Machine learning-based cry diagnostic system for identifying septic newborns. *Journal of Voice*.

Mehta, D. D., Van Stan, J. H., Zañartu, M., Ghassemi, M., Guttag, J. V, Espinoza, V. M., Cortés, J. P., Cheyne, H. A., & Hillman, R. E. (2015). Using ambulatory voice monitoring to investigate common voice disorders: Research update. *Frontiers in Bioengineering and Biotechnology*, *3*, 155.

Menard, S. (2002). *Applied logistic regression analysis* (Issue 106). Sage.

Mielens, J. D., Hoffman, M. R., Ciucci, M. R., McCulloch, T. M., & Jiang, J. J. (2012). *Application of classification models to pharyngeal high-resolution manometry*.

Oleson, J. J., Brown, G. D., & McCreery, R. (2019). The evolution of statistical methods in speech, language, and hearing sciences. *Journal of Speech, Language, and Hearing Research*, *62*(3), 498–506.

Ossewaarde, R., Jonkers, R., Jalvingh, F., & Bastiaanse, R. (2020). Quantifying the Uncertainty of Parameters Measured in Spontaneous Speech of Speakers With Dementia. *Journal of Speech, Language, and Hearing Research*, *63*(7), 2255–2270.

Panayotov, V., Chen, G., Povey, D., & Khudanpur, S. (2015). Librispeech: an asr corpus based on public domain audio books. *2015 IEEE International Conference on Acoustics, Speech and Signal Processing (ICASSP)*, 5206–5210.

Parvandeh, S., Yeh, H.-W., Paulus, M. P., & McKinney, B. A. (2020). Consensus features nested cross-validation. *Bioinformatics*, *36*(10), 3093–3098.

Qayyum, A., Qadir, J., Bilal, M., & Al-Fuqaha, A. (2020). Secure and robust machine learning for healthcare: A survey. *IEEE Reviews in Biomedical Engineering*, *14*, 156–180.

Sagi, O., & Rokach, L. (2018). Ensemble learning: A survey. *Wiley Interdisciplinary Reviews: Data Mining and Knowledge Discovery*, *8*(4), e1249.

Shen, L., Kann, B. H., Taylor, R. A., & Shung, D. L. (2021). The Clinician's Guide to the Machine Learning Galaxy. *Frontiers in Physiology*, *12*, 658583.

Thabtah, F., & Peebles, D. (2020). A new machine learning model based on induction of rules for autism detection. *Health Informatics Journal*, *26*(1), 264–286.

Theodoridis, S., & Koutroumbas, K. (2009). Pattern recognition. In *Elsevier Inc* (fourth).

Tsanas, A., Little, M. A., McSharry, P. E., Spielman, J., & Ramig, L. O. (2012). Novel speech signal processing algorithms for high-accuracy classification of Parkinson's disease. *IEEE Transactions on Biomedical Engineering*, *59*(5), 1264–1271.

Uhm, T., Lee, J. E., Yi, S., Choi, S. W., Oh, S. J., Kong, S. K., Lee, I. W., & Lee, H. M. (2021). Predicting hearing recovery following treatment of idiopathic sudden sensorineural hearing loss with machine learning models. *American Journal of Otolaryngology*, *42*(2), 102858.

Vabalas, A., Gowen, E., Poliakoff, E., & Casson, A. J. (2019). Machine learning algorithm validation with a limited sample size. *PloS One*, *14*(11), e0224365.

Van Stan, J. H., Burns, J., Hron, T., Zeitels, S., Panuganti, B. A., Purnell, P. R., Mehta, D. D., Hillman, R. E., & Ghasemzadeh, H. (2023). Detecting Mild Phonotrauma in Daily Life. *The Laryngoscope*.

Van Stan, J. H., Mehta, D. D., Ortiz, A. J., Burns, J. A., Toles, L. E., Marks, K. L., Vangel, M., Hron, T., Zeitels, S., & Hillman, R. E. (2020). Differences in weeklong ambulatory vocal behavior between female patients with phonotraumatic lesions and matched controls. *Journal of Speech, Language, and Hearing Research*, *63*(2), 372–384.

Varoquaux, G., Raamana, P. R., Engemann, D. A., Hoyos-Idrobo, A., Schwartz, Y., & Thirion, B. (2017). Assessing and tuning brain decoders: cross-validation, caveats, and guidelines. *NeuroImage*, *145*, 166–179.

Vieira, F. G., Venugopalan, S., Premasiri, A. S., McNally, M., Jansen, A., McCloskey, K., Brenner, M. P., & Perrin, S. (2022). A machine-learning based objective measure for ALS disease severity. *NPJ Digital Medicine*, *5*(1), 1–9.

Viering, T., & Loog, M. (2022). The shape of learning curves: a review. *IEEE Transactions on Pattern Analysis and Machine Intelligence*.

Wang, J., Green, J. R., Samal, A., & Yunusova, Y. (2013). Articulatory distinctiveness of vowels and consonants: A data-driven approach. *Speech, Language, and Hearing Research*, *56*, 1539–1551.

Wang, J., Samal, A., Rong, P., & Green, J. R. (2016). An optimal set of flesh points on tongue and lips for speech-movement classification. *Journal of Speech, Language, and Hearing Research*, *59*(1), 15–26.

Wong, P. C. M., Lai, C. M., Chan, P. H. Y., Leung, T. F., Lam, H. S., Feng, G., Maggu, A. R., & Novitskiy, N. (2021). Neural speech encoding in infancy predicts future language and communication difficulties. *American Journal of Speech-Language Pathology*, *30*(5), 2241–2250.

Yousef, A. M., Deliyski, D. D., Zacharias, S. R. C., & Naghibolhosseini, M. (2022). Detection of vocal fold image obstructions in high-speed videoendoscopy during connected speech in adductor spasmodic dysphonia: A convolutional neural networks approach. *Journal of Voice*.

Zhang, Z. (2020). Estimation of vocal fold physiology from voice acoustics using machine learning. *The Journal of the Acoustical Society of America*, *147*(3), EL264–EL270.

Appendix:

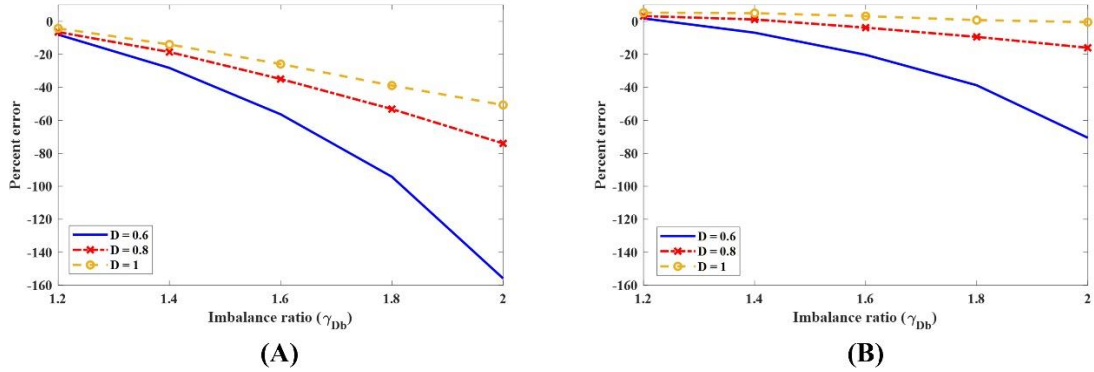

Figure A1. The effect of an unbalanced dataset (e.g., unequal number of patients and controls) on the estimation error of the minimum required sample size when Equations 4–7 were used to estimate (A) the size of the smaller class and (B) an adjusted sample size, where the adjusted sample size was defined as the average of the sample sizes of the positive and negative classes. Please refer to the generalization of the results section for further details.

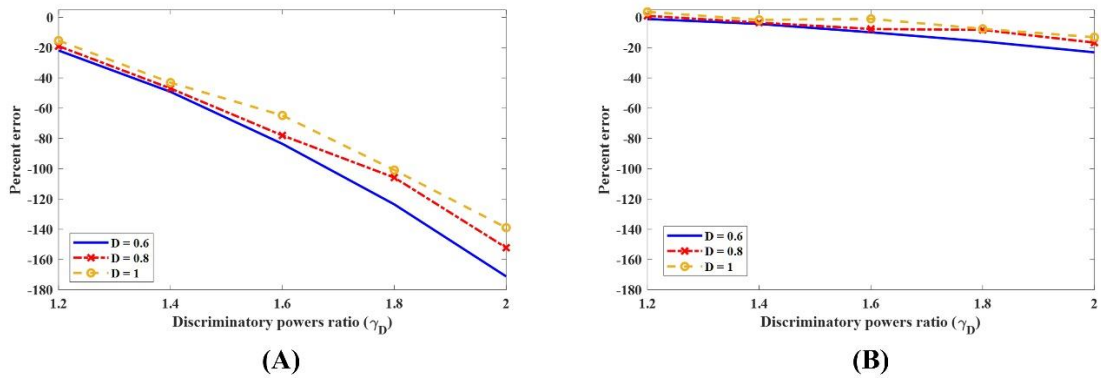

Figure A2. The effect of having two features with unequal discriminative power on the estimation error of the minimum required sample size when Equations 4–7 were used with (A) the smaller effect size ($D_0 = D$) and (B) the average discriminatory power of the two features ($D_0 = D \frac{(\gamma_D+1)}{2}$). Please refer to the generalization of the results section for further details.

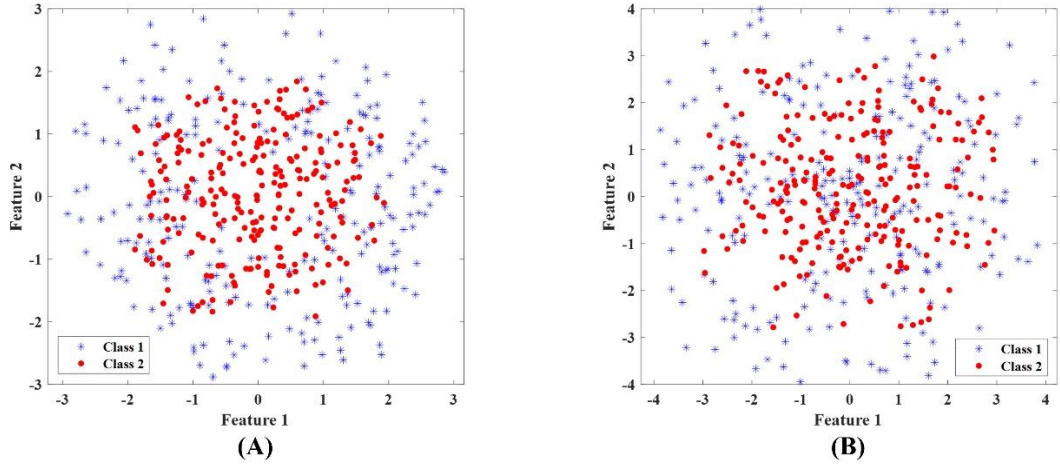

Figure A3. Bull's eye non-linear feature space: (A) with low level of noise (noise level = 2) (B) with high level of noise (noise level = 4). Please refer to the generalization of the results section for further details.

TABLES:

Table A1. Effect of different parameters on the probability of both selected features to be correct ($C_{2,2}$) when $m$ = 10 or 20 features were extracted.

|  |  | Discriminatory power (D), $m$ = 10 | | | | | | | Discriminatory power (D), $m$ = 20 | | | | | | |
|---|---|---|---|---|---|---|---|---|---|---|---|---|---|---|---|
|  |  | 0.4 | 0.5 | 0.6 | 0.7 | 0.8 | 0.9 | 1 | 0.4 | 0.5 | 0.6 | 0.7 | 0.8 | 0.9 | 1 |
| Number of pairs ($n$) | 50 | 17.7 | 27.9 | 40.3 | 50.2 | 60.8 | 68.6 | 75.1 | 9.5 | 17.3 | 27 | 37.4 | 48.7 | 59.4 | 65.5 |
|  | 100 | 38.2 | 51.7 | 66.9 | 78.3 | 85.6 | 90.9 | 94.2 | 23.8 | 40.1 | 55.7 | 68 | 79 | 85.8 | 90.6 |
|  | 150 | 52.7 | 69.3 | 81.4 | 90.3 | 94.7 | 97.2 | 98.4 | 39.5 | 59.5 | 75 | 86.3 | 92.6 | 96.5 | 98 |
|  | 200 | 63.4 | 79.7 | 90.1 | 95.7 | 98.7 | 99.5 | 99.8 | 51.3 | 71.5 | 85.8 | 93.5 | 96 | 98.6 | 99.4 |
|  | 250 | 72.9 | 88.3 | 95.9 | 98.5 | 99.6 | 99.9 | 100 | 63.3 | 83.2 | 92.5 | 96.8 | 99.2 | 99.7 | 99.8 |
|  | 300 | 79.6 | 90.5 | 96.6 | 99.1 | 99.6 | 99.7 | 99.9 | 73.4 | 88.4 | 96.9 | 99.1 | 99.7 | 100 | 100 |
|  | 350 | 84.7 | 94.5 | 98.6 | 99.7 | 99.9 | 100 | 100 | 79 | 92 | 97.5 | 99.4 | 99.8 | 100 | 99.9 |
|  | 400 | 88.1 | 96.1 | 99.1 | 99.8 | 100 | 100 | 100 | 84.1 | 94.9 | 99 | 99.8 | 100 | 100 | 100 |
|  | 450 | 90.3 | 97.3 | 99.6 | 100 | 100 | 100 | 100 | 88.1 | 96.8 | 99.2 | 99.9 | 100 | 100 | 100 |
|  | 500 | 92.6 | 98.6 | 99.9 | 100 | 100 | 100 | 100 | 90.3 | 97.6 | 99.7 | 99.9 | 100 | 100 | 100 |

Table A2. Effect of different parameters on the probability of both selected features to be correct ($C_{2,2}$) when $m$ = 30 or 40 features were extracted.

|  |  | Discriminatory power (D), $m$ = 30 | | | | | | | Discriminatory power (D), $m$ = 40 | | | | | | |
|---|---|---|---|---|---|---|---|---|---|---|---|---|---|---|---|
|  |  | 0.4 | 0.5 | 0.6 | 0.7 | 0.8 | 0.9 | 1 | 0.4 | 0.5 | 0.6 | 0.7 | 0.8 | 0.9 | 1 |
| Number of pairs ($n$) | 50 | 6.3 | 11.9 | 19.7 | 31.5 | 40.9 | 50.7 | 59.7 | 4.8 | 10.3 | 16.6 | 26.5 | 38.2 | 48.2 | 57.5 |
|  | 100 | 19.3 | 35.3 | 52.3 | 67.5 | 77.6 | 85.4 | 90.1 | 15.1 | 31.7 | 46.3 | 60.9 | 72.8 | 81.2 | 87.8 |
|  | 150 | 32.6 | 53.6 | 70.6 | 83.7 | 90.5 | 94.7 | 97.7 | 29.2 | 50.3 | 67.5 | 81 | 89.8 | 94.4 | 97.6 |
|  | 200 | 48.4 | 69.8 | 84.5 | 92.3 | 96.6 | 98.8 | 99.4 | 41.8 | 66.8 | 82.4 | 91.4 | 95.4 | 98.4 | 99.4 |
|  | 250 | 56.8 | 77.5 | 90.6 | 96.3 | 98.7 | 99.5 | 99.9 | 53.3 | 74.1 | 89.3 | 95.4 | 98.3 | 99.5 | 100 |
|  | 300 | 66.1 | 84 | 94.2 | 97.9 | 99.3 | 99.8 | 100 | 63.1 | 81.6 | 93 | 98 | 99.4 | 99.8 | 100 |
|  | 350 | 75.8 | 89.8 | 96.5 | 99.4 | 99.9 | 100 | 100 | 70.8 | 89.1 | 95.4 | 98.8 | 99.6 | 99.8 | 100 |
|  | 400 | 81.2 | 94.1 | 98.7 | 99.8 | 100 | 100 | 100 | 76.1 | 91.2 | 97.8 | 99.4 | 99.9 | 99.9 | 100 |
|  | 450 | 84.8 | 95.7 | 98.9 | 99.7 | 100 | 100 | 100 | 82.9 | 94.9 | 98.8 | 99.7 | 100 | 100 | 100 |
|  | 500 | 86.9 | 96.5 | 99.5 | 100 | 100 | 100 | 100 | 86.8 | 97.3 | 99.6 | 99.9 | 100 | 100 | 100 |

**Description of supplemental files:**

The "Compute_RequiredSampleSize.m" code takes the number of extracted features ($m\_0$), the effect size of the best features ($D\_0$), and the number of selected features ($l\_0$) as the input parameters and computes the required number of pairs for having the statistical power for finding a statistically significant model ($\alpha = 0.05$, $1-\beta = 0.8$). This code can also be used for unbalanced datasets; however, the estimated required sample size ($n_r$) would be the average of the number of negative and positive samples. The sample size of each class can then be computed using the estimated average required sample size and $\gamma_{Db}$. Specifically, the required sample size of the smaller class will be $n_r \times \frac{2}{1+\gamma_{Db}}$ and the required sample size of the larger class will be $n_r \times \frac{2\times\gamma_{Db}}{1+\gamma_{Db}}$. Similarly, this code can also be used for features with different effect sizes; the average of the effect sizes needs to be input as $D\_0$.

The "Compute_RecommendedSampleSize.m" code takes the number of extracted features ($m\_0$), the effect size of the best features ($D\_0$), and the target value of $C_{2,2}$ ($CI\_0$) as the input parameters and computes the recommended number of pairs for achieving the target value of $C_{2,2}$.

The "Compute_NestedModelConfidence.m" code takes the number of extracted features ($m\_0$), the effect size of the best features ($D\_0$), and the number of pairs in a dataset ($n\_0$) as the input parameters and computes the value of $C_{2,2}$.

The "Feature_Selection" toolbox provides an implementation of forward feature selection with four investigated cross-validation approaches. Please, refer to the "Main_Test.m" file for examples of the usage of the code.